\documentclass[10pt,twocolumn,twoside] {IEEEtran}
\def\comment#1{}
\usepackage{graphicx,subfigure,amsfonts,amsmath,amssymb,xcolor}
\usepackage{algorithm,algorithmic}
\usepackage{multirow}
\usepackage[noadjust]{cite}
\def\comment#1{}

\newcommand{\argmin}{\arg\!\min}
\newcommand{\Amat}{{\bf A}}
\newcommand{\Bmat}{{\bf B}}

\newcommand{\Emat}[0]{{{\bf E}}}
\newcommand{\Fmat}[0]{{{\bf F}}}

\newcommand{\Imat}{{\bf I}}

\newcommand{\Umat}{{{\bf U}}}

\newcommand{\Xmat}{{\bf X}}

\newcommand{\gv}[0]{{\boldsymbol{g}}}

\newcommand{\nv}{\boldsymbol{n}}

\newcommand{\qv}[0]{{\boldsymbol{q}}}

\newcommand{\uv}[0]{{\boldsymbol{u}}}
\newcommand{\vv}{\boldsymbol{v}}
\newcommand{\wv}{\boldsymbol{w}}

\newcommand{\xv}{\boldsymbol{x}}
\newcommand{\yv}{\boldsymbol{y}}

\newcommand{\zv}{\boldsymbol{z}}

\newcommand{\Gammamat}[0]{{\boldsymbol{\Gamma}}}

\newcommand{\Sigmamat}{\boldsymbol{\Sigma}}

\newcommand{\Omegamat}{{\boldsymbol{\Omega}}}

\newcommand{\alphav}{\boldsymbol{\alpha}}

\newcommand{\thetav}{\boldsymbol{\theta}}

\newcommand{\muv}{\boldsymbol{\mu}}
\newcommand{\nuv}{{\boldsymbol{\nu}}}
\newcommand{\xiv}{{\boldsymbol{\xi}}}

\newcommand{\rhov}[0]{{\boldsymbol{\rho}} }

\newcommand{\ts}{^{\top}}
\newcommand{\ie}{{\em i.e.}}

\begin{document}
\setlength{\parskip}{.02in}

\title{Compressive Sensing via \\Low-Rank Gaussian Mixture Models}

\author{
\authorblockN{Xin Yuan, Hong Jiang, Gang Huang, Paul A. Wilford} \\
\authorblockA{Bell Labs, Alcatel-Lucent, 600 Montain Avenue, Murray Hill, NJ, 07974, USA}
}

\maketitle

\begin{abstract}
We develop a new compressive sensing (CS) inversion algorithm by utilizing the Gaussian mixture model (GMM).
While the compressive sensing is performed globally on the entire image as implemented in our lensless camera, a low-rank GMM is imposed on the local image patches.
This low-rank GMM is derived via eigenvalue thresholding of the GMM trained on the projection of the measurement data, thus learned {\em in situ}. 
The GMM and the projection of the measurement data are updated iteratively during the reconstruction.
Our GMM algorithm degrades to the piecewise linear estimator (PLE) if each patch is represented by a single Gaussian model. Inspired by this, a low-rank PLE algorithm is also developed for CS inversion, constituting an additional contribution of this paper. 
Extensive results on both simulation data and real data captured by the lensless camera demonstrate the efficacy of the proposed algorithm.
Furthermore, we compare the CS reconstruction results using our algorithm with the JPEG compression. Simulation results demonstrate that when limited bandwidth is available (a small number of measurements), our algorithm can achieve comparable results as JPEG. 
\end{abstract}


\maketitle
\section{Introduction}
\label{Sec:intro}
Compressive sensing~\cite{Donoho06ITT,Candes06ITT,Candes08SPM} (CS) has led to real applications, including the single-pixel camera~\cite{Baraniuk07SPM}, the lensless camera~\cite{Huang13ICIP,Yuan15Lensless}, video compressive sensing~\cite{Patrick13OE,Yuan14CVPR,Yuan13ICIP,Yuan15FiO,Stevens15ASCI}, depth compressive sensing~\cite{Yuan14CVPR,Llull14COSI}, hyperspectral compressive imaging~\cite{Wagadarikar08CASSI,Yuan15JSTSP,Tsai15OL}, polarization compressive sensing~\cite{Tsai15OE}, terahertz imaging~\cite{Chan08APL}, and millimeter wave imaging~\cite{Babacan11ICIP}.
In this paper, we focus on the two-dimensional (2D) image case, though similar technique can be used for videos~\cite{Yang14GMMonline} and other bandwidths. Specifically, we develop our algorithm under the lensless compressive imaging architecture~\cite{Huang13ICIP,Jiang14APSIPA}, which has provided excellent reconstruction images from the compressive measurements using simple and off-the-shelf hardware~\cite{Jiang14APSIPA}. 

Diverse algorithms~\cite{Figueiredo07GPSR,Figueiredo07MM,Yin08bregman,Candes08L1,Tropp07ITT,daubechies2010iteratively} have been developed for compressive sensing recovery, which plays a pivot role in CS, to reconstruct the desired signal from compressive measurements. Sparsity, the key ingredient of CS, has been investigated extensively in these algorithms.
The wavelet transformation~\cite{Bioucas-Dias2007TwIST,Ji08SPT,He09SPT,Figueiredo07GPSR} is generally used since it provides the sparse representation of an image with fast transformations (thus very efficient).
A parallel research is using the total variation (TV) for CS recovery~\cite{Yin08bregman,Li13COA}, which provides good results for piecewise smooth signals.
The aforementioned algorithms do not increase the unknown parameters significantly during the reconstruction, as usually the wavelet coefficients will have similar (or the same) number of the image pixels. 
Recently, researchers have found that by exploiting the sparsity of local patches~\cite{Huang14TIP,Averbuch12SIAM}, better results can be achieved.
In summary, CS recovery algorithms fall into the following three categories:
1) global basis based algorithm, {\em e.g.}, using the wavelet transformation,
2) TV based algorithm, and 3) local basis based algorithm, {\em i.e.}, DCT or dictionary learning or denoising based algorithms.
State-of-the-art CS inversion results have been obtained in~\cite{Dong14TIP,Mertzler14Denoising}, which generally lie within the third category.
In this paper, we propose an alternative inversion algorithm that exploits the low-rank property of image patches.

Generally, the CS recovery is an iterative process in which two steps are performed at each iteration~\cite{Mertzler14Denoising}: $i)$ projecting the measurements to the image level, which can be done by the majorization-minimization (MM) approach~\cite{Figueiredo07MM,Beck09IST}, or the Euclidean projection~\cite{Liao14GAP}, or the alternating direction method of multipliers (ADMM)~\cite{ADMM2011Boyd}, $ii)$ denoising this projected image and updating the recovery of the desired image.
These two steps are performed {\em iteratively} until some criterion is satisfied. 
A general framework is developed in~\cite{Mertzler14Denoising} under the approximate message passing (AMP) framework, in which diverse denoising algorithms~\cite{Dabov07BM3D} can be plugged in.
One key difference of the denoising based CS inversion algorithms compared with the wavelet based CS inversion algorithms is that the former exploits the local sparsity based on (overlapping) patches, as state-of-the-art denoising algorithm is using sparse representation of local patches, {\em e.g.},~\cite{Elad06TIP}. The number of coefficients for the patches (under some basis or dictionary) is usually larger than the image pixel number (because the overlapping patches are used).

\subsection{Contributions}
The proposed algorithm in this paper also lies in the third category mentioned earlier, which is based on the local basis (for image patches). 
Specific contributions of this work can be summarized as below:
\begin{itemize}
	\item We investigate the low-rank property of image patches under the Gaussian mixture model (GMM) framework. Different from the low-rank model investigated in~\cite{Dong14TIP}, which requires patch clustering (block matching) as an additional step, in our algorithm, each patch is modeled by a GMM with different weights corresponding to different Gaussian components, which can be seen as a {\em soft clustering} approach based on these weights.
	Furthermore, these weights are updated in each iteration.
	\item We develop an general framework using ADMM to explore the sparsity (and low-rank property) of patches in order to recover the desired signal.
	\item If each patch is modeled by a single Gaussian component, our GMM degrades to the piecewise linear estimator (PLE)~\cite{Yu11SPT,Yu12IPT}. Therefore, a low-rank PLE algorithm is also proposed for CS recovery.
	\item We conduct experiments using our proposed algorithm and other leading algorithms on the real data captured by our lensless camera. This verifies real applications of each algorithm.
\end{itemize}

\subsection{Organization of This Paper}
We start with the derivation of an ADMM formulation for CS recovery to investigate the sparsity of local overlapping patches in Section~\ref{Sec:ADMM_slope}. The proposed low-rank GMM algorithm is developed in Section~\ref{Sec:GMM} and the joint reconstruction algorithm is summarized in Section~\ref{Sec:Joint}.
Extensive results on both simulation and real data are reported in Sections~\ref{Sec:simResults}-\ref{Sec:realResults}. Section~\ref{Sec:Con} concludes the paper.


\section{CS Inversion via Exploiting Sparsity of Patches}
\label{Sec:ADMM_slope}
Under the CS framework, the problem we are solving can be formulated as:
\begin{eqnarray}
&\min &\frac{1}{2}\|\yv - \Amat\xv\|^2_2 + \lambda \|\zv\|_1,\\
&{\text{s.t.}} & \xv = \Bmat \zv, \label{eq:xBz}
\end{eqnarray}
where $\Amat\in {\mathbb R}^{M\times N}$ is the sensing matrix, $\xv\in {\mathbb R}^N$ is the desired signal, $\zv$ is the coefficients which are sparse under the basis $\Bmat$. From $\zv$, we can recover $\xv$ easily via $\Bmat$, which can be {\em known a priori}, ({\em e.g.}, a wavelet or DCT basis) or learned based on $\xv$ during the reconstruction.

Considering the image case investigated here, let $\xv$ denote the vectorized image and the sparse representation, $\zv$, is now modeled on image patches. Therefore, (\ref{eq:xBz}) can be reformulated as:
\begin{equation}
{\bf R} \xv = \Bmat \zv,
\end{equation}
where $\bf{R}$ denotes the patch extraction and vectorization operation and considering each patch, we have
\begin{equation}
{\bf R}_i \xv = \Bmat \zv_i
\end{equation}
with $i$ indexes the patches.

The problem can be reformulated as:
\begin{eqnarray}\label{eq:P1}
&\min &\frac{1}{2}\|\yv - \Amat\xv\|^2_2 + \lambda \|\zv\|_1,\\
&{\text{s.t.}} & {\bf R}_i \xv = \Bmat \zv_i.
\end{eqnarray}
Note this $\Bmat$ is shared for all patches for current discussion, and in the following analysis, similar patches can be grouped together~\cite{Dong14TIP} and each group can have its own $\Bmat$.

Next, we develop an ADMM~\cite{ADMM2011Boyd} formulation of \eqref{eq:P1} to solve the problem, which will also be used in our GMM formulation in Section~\ref{Sec:Joint}. 
Introducing Lagrange multipliers $\rhov$ and  parameter $\eta$ in (\ref{eq:P1}) results in the objective function
\begin{align}
L(\xv, \zv, \lambda, \rho) &= \frac{1}{2}\|\yv - \Amat\xv\|^2_2 + \lambda \|\zv\|_1 + \rhov^{\top} \sum_i({\bf R}_i \xv - \Bmat \zv_i) \nonumber\\
& \quad + \frac{\eta}{2}\sum_i \|{\bf R}_i \xv - \Bmat \zv_i\|_2^2
\end{align}
Define $\uv = (1/\eta) \rhov$,
\begin{eqnarray}
L(\xv, \zv, \uv,\lambda) &=& \frac{1}{2}\|\yv - \Amat\xv\|^2_2 + \lambda \|\zv\|_1 \nonumber\\
&& + \frac{\eta}{2}\sum_i \|{\bf R}_i \xv - \Bmat \zv_i + \uv\|_2^2 + {\rm const}.
\end{eqnarray}
The ADMM cyclically solves the following 3 sub-problems:
\begin{align}
\xv^{t+1} &:= \arg \min_{\xv} (\frac{1}{2}\|\yv - \Amat\xv\|^2_2 + \frac{\eta}{2}\sum_i \|{\bf R}_i \xv - \Bmat \zv^{t}_i + \uv^t\|_2^2 ) 
\label{eq:x_k+1}\\
\zv^{t+1} &:= \arg \min_{\zv} (\lambda \|\zv\|_1 + \frac{\eta}{2}\sum_i \|{\bf R}_i \xv^{t+1} - \Bmat \zv_i + \uv^t\|_2^2) 
\label{eq:z_k+1}\\
\uv^{t+1}&:= \uv^{t} + \sum_i({\bf R}_i \xv^{t+1} - \Bmat \zv_i^{t+1})
\label{eq:u_k+1}
\end{align} 
where $t$ denotes the iteration index.

Equation (\ref{eq:x_k+1}) is a quadratic optimization problem and can be simplified to
\begin{equation}
(\Amat^{\top} \Amat + \eta\sum_i {\bf R}_i^{\top} {\bf R}_i) {\xv} = 
\Amat^{\top} \yv + \eta \sum_i({\bf R}_i^{\top} \Bmat \zv_i - {\bf R}_i^{\top} \uv^{t}),
\end{equation}
which admits the closed-form solution,
\begin{equation}\label{eq:xinv}
{\xv} = 
(\Amat^{\top} \Amat + \eta\sum_i {\bf R}_i^{\top} {\bf R}_i)^{-1}[\Amat^{\top} \yv + \eta \sum_i({\bf R}_i^{\top} \Bmat \zv_i - {\bf R}_i^{\top} \uv^{t})].
\end{equation}
However, the dimension of $(\Amat^{\top} \Amat + \eta\sum_i {\bf R}_i^{\top} {\bf R}_i) $ is large (the pixels of the desired image), requiring a high computational workload.
Alternatively, since $(\Amat^{\top} \Amat + \eta\sum_i {\bf R}_i^{\top} {\bf R}_i) $ is invertible, the matrix inversion formula can be used to reduce the computational workload.
As ${\bf R}_i$ is used to extract $i$-th patch from an image,
$\sum_i{\bf R}_i^{\top} {\bf R}_i$ is a diagonal matrix 
\begin{equation}
\tilde{\bf R}\stackrel{\rm def}{=}\sum_i{\bf R}_i^{\top} {\bf R}_i={\rm diag}(r_1,\dots,r_N).
\end{equation} 
Each of the diagonal entries corresponds to an image pixel location and its value is the number of overlapping patches that cover that pixel. 
Therefore, $\tilde{\bf R}^{-1}={\rm diag}(r_1^{-1},\dots,r_N^{-1})$ and 
\begin{align}
&(\Amat^{\top} \Amat + \eta\tilde{\bf R})^{-1} = \eta^{-1}\tilde{\bf R}^{-1} \nonumber\\
&- \eta^{-1}\tilde{\bf R}^{-1}\Amat^{\top} (\Imat + \Amat \eta^{-1}\tilde{\bf R}^{-1}\Amat^{\top})^{-1}\Amat\eta^{-1} \tilde{\bf R}^{-1}.
\end{align}
However, this is not necessarily easy to calculate though $\tilde{\bf R}$ can be pre-computed. $\Amat\tilde{\bf R}^{-1}\Amat^{\top}$ needs to be saved for computation.
Alternatively, (\ref{eq:xinv}) can be solved by the conjugate gradient algorithm~\cite{Jiang15TSP}.


To mitigate this problem, we apply the ADMM again on (\ref{eq:P1}) and introduce another auxiliary variable $\wv$,  leading to the following optimization problem:
\begin{align}\label{eq:P2}
\min &~~\frac{1}{2}\|\yv - \Amat\xv\|^2_2 + \lambda \|\zv\|_1 + \frac{\eta}{2}\sum_i \|{\bf R}_i \wv - \Bmat \zv_i\|_2^2,\\
{\text{s.t.}} &~~ \xv =  \wv.
\end{align}
Following this,
\begin{align}
(\xv,\wv,\zv,\alphav)&=\underset{\xv,\wv,\zv,\alphav}{\argmin}  \frac{1}{2}\|\yv - \Amat\xv\|^2_2  +  \frac{\eta}{2}\sum_i \|{\bf R}_i \wv - \Bmat \zv_i\|_2^2  \nonumber\\
&+ \lambda \|\zv\|_1+ \alphav^{\top}(\xv-\wv) + \frac{\beta}{2}\|\xv-\wv\|_2^2
\end{align}
which can be simplified to (by setting $\vv = \alphav/\beta $):
\begin{align}\label{eq:min3}
(\xv,\wv,\zv,\vv)&=\underset{\xv,\wv,\zv,\vv}{\argmin}  \frac{1}{2} \|\yv - \Amat\xv\|^2_2  +  \frac{\eta}{2}\sum_i \|{\bf R}_i \wv - \Bmat \zv_i\|_2^2  \nonumber\\
&~~+ \lambda \|\zv\|_1 + \frac{\beta}{2}\|\xv-\wv + \vv\|_2^2 + {\rm const}
\end{align}
The optimization of (\ref{eq:min3}) consists of the following iterations:
\begin{align}
\xv^{t+1} &:= \arg \min_{\xv} \frac{1}{2}\|\yv - \Amat\xv\|^2_2 +\frac{\beta}{2}\|\xv-\wv^k + \vv^t\|_2^2,
\label{eq:x3_k+1}\\
\wv^{t+1}&:= \arg\min_{\wv}\frac{\eta}{2}\sum_i \|{\bf R}_i \wv - \Bmat \zv^t_i\|_2^2\nonumber\\ &\qquad \qquad \qquad+ \frac{\beta}{2}\|\xv^{t+1}-\wv + \vv^t\|_2^2, \label{eq:w3_k+1}\\
\zv^{t+1} &:= \arg \min_{\zv} \lambda \|\zv\|_1 + \eta\sum_i \|{\bf R}_i \wv^{t+1} - \Bmat \zv_i \|_2^2,
\label{eq:z3_k+1}\\
\boldsymbol{v}^{t+1}&:= \boldsymbol{\vv}^{t} + (\xv^{t+1} - \wv^{t+1}).
\label{eq:v3_k+1}
\end{align}
%
For fixed \{$\wv^t, \vv^t$\}, $\xv^{t+1}$ admits the following closed-form solution:
\begin{align}
\xv^{t+1} &= (\Amat^{\top}\Amat + \beta \Imat)^{-1}[\Amat^{\top}\yv + \beta (\wv^t - \vv^t)],
\end{align}
which can be simplified to
\begin{align}
\xv^{t+1} &= (\beta^{-1}{\Imat}-\beta^{-1}\Amat^{\top}(\Imat + \Amat\beta^{-1}\Amat^{\top})^{-1}\Amat\beta^{-1})\nonumber\\
&\times[\Amat^{\top}\yv + \beta (\wv^t - \vv^t)],
\end{align}
For the case considered in our work (as implemented in the lensless camera~\cite{Huang13ICIP}), $\Amat$ is the permuted Hadamard matrix and thus $\Amat\Amat^{\top}$ is an identity matrix:
\begin{align}\label{eq:xp3_k+1}
\xv^{t+1} &= \left(\beta^{-1}{\Imat} - \frac{\Amat^{\top} \Amat}{(\beta+1)\beta}\right)[\Amat^{\top}\yv + \beta (\wv^t - \vv^t)]\nonumber\\
& = \frac{\Amat^{\top}\yv}{\beta+1} + (\wv^t - \vv^t) - \frac{\Amat^{\top}\Amat(\wv^t - \vv^t)}{\beta +1} \nonumber \\
& = (\wv^t - \vv^t) + \frac{\Amat^{\top}(\yv - \Amat(\wv^t - \vv^t))}{\beta+1}.
\end{align}
Similarly, for fixed \{$\xv^{t+1}, \vv^t, \zv^t$\}, $\wv^{t+1}$ admits the following closed-form solution:
\begin{align}\label{eq:wp3_k+1}
 \wv^{t+1} &= \left(\eta\sum_i{\bf R}_i^{\top} {\bf R}_i + \beta \Imat\right)^{-1}\nonumber\\
 &\quad\times\left[\beta(\xv^{t+1}+ \vv^t) + \eta\sum_i{\bf R}^{\top}_i {\bf B} \zv_i^t\right]
\end{align}
Recall that $\sum_i{\bf R}_i^{\top} {\bf R}_i$ is a diagonal matrix $\tilde{\bf R}\stackrel{\rm def}{=}{\rm diag}(r_1,\dots,r_N)$, thus $\wv$ can be computed element wise via
\begin{align}\label{eq:wnk+1}
w_n^{t+1}&= \frac{\left[\beta(\xv^{t+1}+ \vv^t) + \eta\sum_i{\bf R}^{\top}_i {\bf B} \zv_i^t\right]_n}{\eta r_n + \beta}
\end{align}
where $[\cdot]_n$ denotes the $n$-th entry of the vector inside $[~]$.

Similar to (\ref{eq:z_k+1}), (\ref{eq:z3_k+1}) can be considered as a dictionary learning model, where $\Bmat$ is the dictionary. If the orthonormal transformation is used ({\em e.g.}, the DCT), $\zv$ can be solved by the shrinkage thresholding operation~\cite{Figueiredo07MM,Bioucas-Dias2007TwIST}.

Since the key of this algorithm is to investigate the sparsity of the local overlapping patches, we term this framework as SLOPE (Shrinkage of Local Overlapping Patches Estimator), where the `local' stands for the local basis rather than the global basis such as wavelet.
The ADMM-SLOPE is summarized in Algorithm~\ref{algo:slope}.

While good results have been obtained using similar approaches~\cite{Huang14TIP,Dong14TIP} as in Algorithm~\ref{algo:slope}, in the next section, we develop a low-rank GMM framework imposed on the patches and a full formulation of the proposed algorithm is presented in Section~\ref{Sec:Joint}.

\begin{center}
	\begin{algorithm}[htbp]
		\caption{ADMM-SLOPE}
		\begin{algorithmic}[1]
			\REQUIRE Measurements ${\yv}$, sensing matrix $\Amat$, \{$\beta$, $\eta$, $\lambda$\}.
			\STATE Initial $\xv, \wv, \vv$ to all 0.
			\FOR{$t=1$ \TO Max-Iter }
			\STATE Update $\xv$ by Eq. (\ref{eq:xp3_k+1}).
			\STATE Update $\wv$ by Eq. (\ref{eq:wp3_k+1}).
			\STATE Update $\zv$ by shrinkage operator.
			\STATE Update $\vv$ by Eq. (\ref{eq:v3_k+1}).
			\ENDFOR
		\end{algorithmic}
		\label{algo:slope}
	\end{algorithm}
\end{center}

\section{The Gaussian Mixture Model}
\label{Sec:GMM}
The Gaussian mixture model (GMM) has been re-recognized as an advanced dictionary learning approach and has achieved excellent results in image processing~\cite{Yu11SPT,Yu12IPT} and video compressive sensing~\cite{Yang14GMM,Yang14GMMonline}.
Recall the image patches  $\Xmat \in{\mathbb R}^{P \times N_p}$ extracted from the 2D image, where the patch size is $\sqrt{P} \times \sqrt{P}$ and there are in total $N_p$ patches. For $i$-th patch $\xv_i$, it is modeled by a GMM with $K$ Gaussians~\cite{Chen10SPT}:
\begin{equation}\label{eq:xiGMM}
\xv_i \sim \sum_{k=1}^K \pi_k {\cal N}(\muv_k, \Sigmamat_k)
\end{equation} 
where $\{\muv_k, \Sigmamat_k\}_{k=1}^K$ represent the mean and covariance matrix of $k$-the Gaussian, and $\{\pi_k\}_{k=1}^K$ denotes the weights of these Gaussian component. 

In this paper, we further impose the GMM is low-rank and 
now the model in (\ref{eq:xBz}) becomes (\ref{eq:xiGMM}) and the problem to be solved becomes
\begin{eqnarray}
&\min &\frac{1}{2}\|\yv - \Amat\xv\|^2_2 \\
&{\text{s.t.}} & \xv_i \sim \sum_{k=1}^K \pi_k {\cal N}(\tilde{\muv}_k, \tilde{\Sigmamat}_k)
\end{eqnarray}
where $\xv_i$ denotes $i$-th patch from $\xv$, which is an vectorized image. $\{\tilde{\muv}_k, \tilde{\Sigmamat}_k\}_{k=1}^K$ symbolize the low-rank GMM.

The following problem is to estimate this low-rank GMM. Recalling  Section~\ref{Sec:intro},  we review that the CS recovery is an iterative two-step procedure. In each iteration, one can get an estimate from the projection of the measurements (details discussed in Section~\ref{Sec:Joint}). We hereby learn a (full rank) GMM from this estimate and then proposing the eigenvalue thresholding approach to derive the low-rank GMM based on this full rank GMM. 

\subsection{Low-Rank GMM}
In the following, we provide a motivation for the next step in the algorithm.
The random vector $\xv_i$ in (\ref{eq:xiGMM}) (modeled as a GMM) can be written as, dropping the subscript $i$ for simplicity,
\begin{equation}
\xv = \sum_{k=1}^K \pi_k \gv_k,
\end{equation}
where each $\gv_k$ is a random vector of multivariate normal distribution, given by
\begin{equation}\label{eq:ziNormal}
\gv_k \sim {\cal N}(\muv_k, \Sigmamat_k).
\end{equation}
The random vector $\gv_k$ can be decomposed into independent random variables of normal distribution~\cite{Gut2009,Yang14GMM} as follows
\begin{equation}\label{eq:ziDecomp}
\gv_k = \Fmat_k \qv_k + \muv_k, 
\end{equation}
where $ \Fmat_k \in {\mathbb R}^{P\times \gamma_k}, \qv_k \in {\mathbb R}^{\gamma_k}$,
$\gamma_k$ is the rank of $\Sigmamat_k$, and $\qv_k$ is a vector whose $\gamma_k$ components are independent random variables. 

In order to reduce noise, we use a model in which $\gv_k$ has a small number of independent random components, \ie, we require $\gamma_k$ to be small. This is equivalent to requiring $\Sigmamat_k$ have a reduced rank. More specifically, introducing the parameter $\tau_k$, we solve the following minimization problem \cite{Cai10SVT}:
\begin{align}\label{eq:SigmaLowRank}
\tilde{\Sigmamat}_k = \arg\min_{\Gammamat} \{{\frac12\|\Gammamat-\Sigmamat_k||^2_F} + \tau_k  \|\Gammamat\|_*\},
\quad \forall k=1,\cdots,K.
\end{align}
where $\|\cdot\|_F$ is the Frobenious norm, and $\|\cdot\|_*$  is the nuclear norm (sum of the singular values). It is shown in \cite{Cai10SVT} that the solution to (\ref{eq:SigmaLowRank}) can be readily obtained by a shrinkage on the singular values (which are similar to the eigenvalues) of  $\Sigmamat_k$. 
Specifically, consider
\begin{eqnarray} \label{eq:LR_siguu}
\Sigmamat_k &=& \Umat_k \Lambda_k \Umat_k\ts, \\
\Lambda_k &=& [\lambda_1, \dots, \lambda_P].
\end{eqnarray}
We impose that $\gamma_k <P$ via
\begin{eqnarray}
\tilde{\Lambda}_k &=& [\tilde{\lambda}_1, \dots, \tilde{\lambda}_{r_k}, {\bf 0}],\\
\tilde{\lambda}_i &=& \max(\lambda_i-\lambda_{\gamma_k +1},  0), \quad \forall i = 1,\dots,P.
\end{eqnarray}
And we term this as the eigenvalue thresholding (EVT).
Following this, $\tilde{\Sigmamat}_k$ is obtained by
\begin{equation}\label{eq:LR_sig}
\tilde{\Sigmamat}_k = \Umat_k \tilde{\Lambda}_k \Umat_k\ts.
\end{equation}

We further define
\begin{equation}\label{eq:mut}
\tilde{\muv}_k = \muv_k, \quad \forall k=1,\cdots,K.
\end{equation}
Next, we define a new random vector
\begin{equation}\label{eq:tdxiGMM}
\tilde{\xv}_i \sim \sum_{k=1}^K {\pi}_k {\cal N}(\tilde{\muv}_k, \tilde{\Sigmamat}_k),
\end{equation} 
which is modeled by a low-rank GMM, parameterized by $\{{\pi}_k, \tilde{\muv}_k, \tilde{\Sigmamat}_k\}_{k=1}^K$.

\subsection{Update Estimate via the Low-Rank GMM}
Given an estimated image ${\hat \xv}$, the GMM in (\ref{eq:xiGMM}) can be learned via the Expectation-Maximization (EM) algorithm~\cite{Chen10SPT,Yang14GMM} based on overlapping patches. 
Then for each Gaussian component, we adopt the EVT to the covariance matrix to obtain the low-rank GMM $\{\tilde{\muv}_k, \tilde{\Sigmamat}_k\}_{k=1}^K$. Following this, the estimated image $\hat{\xv}$ or patches $\xv_i$ can be updated via this low-rank GMM, to $\hat{\tilde \xv}$ or ${\tilde \xv}_i$.
Dropping the subscript $i$, given ${\hat\xv}$, the conditional distribution for $\hat{\tilde \xv}$ maybe evaluated as
\begin{eqnarray}\label{eq:postxhat}
p(\hat{\tilde \xv}|{\hat\xv}) &=& \frac{p(\hat{\tilde \xv})p(\hat{\xv}|{\hat {\tilde \xv}})}{\int p(\hat{\tilde \xv}) p(\hat \xv |{\hat {\tilde{\xv}}}) d{\hat{\tilde \xv}}}
\end{eqnarray}
Since ${\hat {\tilde{\xv}}}$ is a low-rank version of $\hat {\xv}$, we assume:
\begin{eqnarray}
\hat {\xv} &=& \hat {\tilde{\xv}} + \nv,
\end{eqnarray}
where $\nv\sim {\cal N}(0, \Emat)$ is modeled as an additive Gaussian noise, thus
\begin{eqnarray}\label{eq:xxtilde}
p(\hat{\xv}|{\hat {\tilde \xv}})&\sim& {\cal N}({\hat {\tilde \xv}}, \Emat) 
\end{eqnarray} 
Plugging (\ref{eq:xxtilde}) into (\ref{eq:postxhat}), we have
\begin{eqnarray}
p(\hat{\tilde \xv}|{\hat\xv}) &=& \frac{p(\hat{\tilde \xv})p(\hat{\xv}|{\hat {\tilde \xv}})}{\int p(\hat{\tilde \xv}) p(\hat \xv |{\hat {\tilde{\xv}}}) d{\hat{\tilde \xv}}} \\
&=& \frac{\sum_{k=1}^K \pi_k{\cal N}(\tilde{\muv}_k, \tilde{\Sigmamat}_k) \times {\cal N}({\hat {\tilde \xv}}, \Emat)}{\int \sum_{l=1}^K \pi_l{\cal N}(\tilde{\muv}_l, \tilde{\Sigmamat}_l) \times {\cal N}({\hat {\tilde \xv}}, \Emat) d{\hat{\tilde \xv}}} \\
&=& \sum_{k=1}^K \phi_k {\cal N}(\hat{\tilde \xv}; \nuv_k, \Omegamat_k) \label{eq:xv_pdf}
\end{eqnarray}
which is an analytical solution with~\cite{Chen10SPT}
\begin{eqnarray}
\phi_k &=& \frac{\pi_k {\cal N}({\hat \xv}; {\tilde \muv_k},\Emat + \tilde{\Sigmamat}_k)}{\sum_{l=1}^K\pi_l {\cal N}({\hat \xv; {\tilde \muv_k}},\Emat + \tilde{\Sigmamat}_k)}, \\
\Omegamat_k &=& (\Emat^{-1} + \tilde{\Sigmamat}_k^{-1})^{-1}\nonumber\\
&=& \tilde{\Sigmamat}_k - \tilde{\Sigmamat}_k (\Emat + \tilde{\Sigmamat}_k)^{-1} \tilde{\Sigmamat}_k, \\
\nuv_k &=& \Omegamat_k (\Emat^{-1} \hat{{\xv}} + \tilde{\Sigmamat}_k^{-1} \tilde{\muv}_k) \nonumber\\
&=& \tilde{\Sigmamat}_k(\Emat + \tilde{\Sigmamat}_k)^{-1} ({\hat{\xv} - \tilde{\muv}_k}) + \tilde{\muv}_k.
\end{eqnarray}
Note that $\tilde{\Sigmamat}_k$ is low-rank obtained via EVT from $\Sigmamat_k$, but by adding $\Emat$ ($=\sigma^2 \Imat_P$), $(\Emat + \tilde{\Sigmamat}_k)$ is invertible.
While (\ref{eq:xv_pdf}) provides a posterior distribution for $\hat{\tilde{\xv}}$, we obtain the point estimate of $\hat{\tilde \xv}$ via the posterior mean:
\begin{eqnarray}\label{eq:txhatmean}
{\mathbb E}[\hat{\tilde \xv}]&=& \sum_{k=1}^K \phi_k \nuv_k
\end{eqnarray}
which is a closed-form solution.

The procedure of learning and updating the GMM can be summarized as below:
\begin{itemize}
	\item Step 1:
	Lean a GMM (not low-rank) $\{\pi_k, \muv_k, \Sigmamat_k\}_{k=1}^K$  via EM from an estimate of $\hat{\xv}$, which can be obtained from IST, GAP or ADMM described below in Section~\ref{Sec:Joint}.
	\item Step 2:
	For each Gaussian component, derive the low-rank version $\{\tilde{\muv}_k, \tilde{\Sigmamat}_k\}_{k=1}^K$ by eigenvalue value thresholding via (\ref{eq:LR_siguu})-(\ref{eq:mut}).
	\item Step 3:
	Update the estimate of the image by $\hat{\tilde \xv}$ using (\ref{eq:xv_pdf})-(\ref{eq:txhatmean}).
\end{itemize}

\subsection{Degrade to the Piecewise Linear Estimator}
The piecewise linear estimator (PLE) proposed in~\cite{Yu12IPT} has demonstrated excellent performance on diverse image processing tasks.
If each patch is considered drawn from a single Gaussian distribution, our GMM degrades to the PLE and the weights $\pi_k$ (or $\phi_k$) are not required.
Furthermore, the update equation of $\hat{\tilde{\xv}}$ will become the Winer filter.
The MAP-EM procedure proposed in~\cite{Yu12IPT} can still be used to determine which Gaussian each patch lies in and to estimate the denoising version of the each patch.
However, this method has been shown that it is very sensitive to the initialization and selecting $K$ is critical to the performance of the method. We compare our proposed algorithm with PLE by experiments in Section~\ref{Sec:GMM_PLE}.

It is worthing nothing that, even using PLE, the eigenvalue thresholding method used to obtain the low-rank Gaussian model is first proposed in this paper. 
In this case, the PLE model is very similar to the NLR-CS~\cite{Dong14TIP}, where the low-rank is imposed on each cluster of patches, while in the PLE, the low-rank is imposed on the patches belonging to the same Gaussian component; this can also be seen as a cluster.

Next, we review the MAP-EM algorithm proposed in~\cite{Yu12IPT} and adopt it to the current context.
In the E-step, assuming that the estimates of the {\em low-rank} Gaussian parameters $\{\tilde{\muv}_k, \tilde{\Sigmamat}_k\}_{k=1}^K$ are known, (following the previous M-step), for each patch, one calculates the MAP estimates $\thetav_i^{k}$ of all the Gaussian models and selects the best Gaussian model $\tilde{k}_i$ to obtain the estimate of the patch $\tilde{\xv}_i = \thetav_i^{\tilde{k}_i}$.
In the M-step, assuming that the Gaussian models selection $\tilde{k}_i$ and the signal estimate $\tilde{\xv}_i, \forall i$, are known (following the previous E-step), one updates the Gaussian models $\{\tilde{\muv}_k, \tilde{\Sigmamat}_k\}_{k=1}^K$ and then impose them to be low-rank.
\begin{itemize}
	\item {\em E-step: Signal Estimation and Model Selection:}
	
	For each image patch $i$, the signal estimation and the model selection are calculated to maximize the log {\em a posteriori} probability $\log p(\tilde{\xv}_i|\xv_i)$:
	\begin{equation}
	(\tilde{\xv}_i, \tilde{k}_i) = \arg \max_{\thetav,k}\log p(\thetav| \xv_i, \tilde{\muv}_k, \tilde{\Sigmamat}_k)
	\end{equation}
	Recall that we consider $\tilde{\xv}$ is a low-rank version of $\xv_i$ and
	\begin{equation}
	\xv_i = \tilde{\xv}_i + \nv, \quad \nv\sim {\cal N}(0, \sigma^2\Imat_P)
	\end{equation}
	Therefore:
	\begin{align}
	(\tilde{\xv}_i, \tilde{k}_i) &= \arg \max_{\thetav,k}\left(\log p(\xv_i|\thetav,\sigma^2\Imat_p) \right. \nonumber\\
	&\qquad \qquad \qquad \left.+ \log p (\thetav|\tilde{\muv}_k, \tilde{\Sigmamat}_k)\right) \\
	&= \arg\min_{\thetav,k}\left(\sigma^{-2}\|\xv_i-\thetav\|^2 + 0.5 \log |\tilde{\Sigmamat}_k|\right.\nonumber\\
	&\qquad \qquad \left.+ (\thetav-\tilde{\muv}_k)
	\ts\tilde{\Sigmamat}^{-1}_k(\thetav-\tilde{\muv}_k) \right)
	\end{align}
	This maximization is first calculated over $\thetav$ and then over $k$. Given a prior Gaussian signal model $\thetav\sim {\cal N}(\tilde{\muv}_k, \tilde{\Sigmamat}_k)$, $\thetav$ can be estimated by the posteriori mean
	\begin{eqnarray} \label{eq:txv_k}
	\thetav_i^k &=& \tilde{\Sigmamat}_k(\tilde{\Sigmamat}_k + \sigma^2 \Imat_P)^{-1} \xv_i
	\end{eqnarray}
	The best Gaussian model $\tilde{k}_i$ that generates the maximum MAP probability among all the models is then selected with the estimated $\tilde{\xv}_i^k$
	\begin{eqnarray}
	\tilde{k}_i &=& \arg\min_k\left(\sigma^{-2}\|\xv_i-\thetav_i^k\|^2 + 0.5 \log |\tilde{\Sigmamat}_k| \right.\nonumber\\
	&&\qquad \qquad \left.  + (\thetav_i^k-\tilde{\muv}_k)
	\ts\tilde{\Sigmamat}^{-1}_k(\thetav_i^k-\tilde{\muv}_k) \right) \label{eq:t_k}
	\end{eqnarray}
	The signal estimate is obtained by plugging in the best model $\tilde{k}_i$ in the MAP estimate
	\begin{equation}
	\tilde{\xv}_i = \thetav_i^{\tilde{k}_i.} \label{eq:xv_i}
	\end{equation}
	
	\item {\em M-step: Model Estimation:}
	
	In the M-step, the Gaussian model selection $\tilde{k}_i$ and the signal estimate ${\xv}_i$ of all the patches are assumed to be know (derived from the IST, GAP or ADMM as shown in Section~\ref{Sec:Joint}).
	The parameters of each Gaussian model are estimated with the maximum-likelihood (ML) estimate using all the patches in the same Gaussian model:
	\begin{align}
	({\muv}_k, {\Sigmamat}_k)&= \arg \max_{\xiv_k, \Omegamat_k}\log p \left(\{\tilde{\xv}_i\}_{i\in {\cal C}_k}|\xiv_k, \Omegamat_k\right) \label{eq:plemu_sig}
	\end{align}
	where ${\cal C}_k$ denotes the ensemble of the patch indices $i$ that are assigned to the $k$-th Gaussian model and
	\begin{eqnarray}
	{\muv}_k &=& \frac{1}{|{\cal C}_k|} \sum_{i \in{\cal C}_k} \tilde{\xv}_i, \label{eq:plemu}\\
	{\Sigmamat}_k &=& \frac{1}{|{\cal C}_k|} \sum_{i \in{\cal C}_k} (\tilde{\xv}_i-{\muv}_k)(\tilde{\xv}_i-{\muv}_k)\ts. \label{eq:plesig}
	\end{eqnarray}
\end{itemize}
Return to the low-rank model proposed in this paper.
The E-step is same as above and one more step is added in the M-step. The full rank (not low-rank) Gaussian models are first estimated via (\ref{eq:plemu_sig})-(\ref{eq:plesig}), and then each Gaussian model is imposed to be low-rank by thresholding the eigenvalues via (\ref{eq:LR_siguu})-(\ref{eq:LR_sig}).
The new low-rank PLE algorithm can be summarized into the following 3 steps:
\begin{itemize}
	\item Step 1:
	Signal estimation and model selection by (\ref{eq:txv_k})-(\ref{eq:xv_i}).
	\item Step 2:
	Model update for the (non low-rank) Gaussian models $\{\muv_k,\Sigmamat_k\}_{k=1}^K$ by (\ref{eq:plemu_sig})-(\ref{eq:plesig}).
	\item Step 3:
	Estimate the low-rank Gaussian models $\{\tilde{\muv}_k,\tilde{\Sigmamat_k}\}_{k=1}^K$ from  $\{\muv_k,\Sigmamat_k\}_{k=1}^K$ via (\ref{eq:LR_siguu})-(\ref{eq:mut}).
\end{itemize}

\section{The Joint Reconstruction Algorithm}
\label{Sec:Joint}
Section~\ref{Sec:GMM} presents an algorithm to obtain a better estimate (or a denoised version) of the signal $\xv$ given an initial estimate utilizing the low-rank GMM. In this section, the GMM will be wrapped into our joint reconstruction algorithm by different update methods to get the initial estimate, which can be considered as projecting the measurement $\yv$ to the image plane $\xv$.
This is obtained by minimizing the following objection function
\begin{equation} \label{eq:Jx}
J({\bf x}) = \|{\bf y - Ax}\|_2^2.
\end{equation}
Diverse algorithms have been proposed and we review two of them below and develop an ADMM formulation in Section~\ref{Sec:ADMM}.
Other approaches, for example, the TwIST~\cite{Bioucas-Dias2007TwIST} can also be used.

\subsection{Iterative Shrinkage Thresholding}
\label{Sec:IST}
By using the majorization-minimization approach~\cite{Figueiredo07MM} to minimize $J(\xv)$, we can avoid solving a system of linear equations. At each iteration $t$ of the MM approach, we should find a function $G_t(\xv)$ that coincides with $J(\xv)$ at $\xv^t$ but otherwise upper-bounds $J(\xv)$. We should choose
a majorizer $G_t(\xv)$ which can be minimized more easily (without having to solve a system of equations).
The $G_t(\xv)$ is defined as
\begin{equation}
G_t({\xv}) = \|{\xv - \Amat\xv}\|_2^2 + (\xv -\xv^t)^{\top}(\zeta \Imat - \Amat^{\top} \Amat)(\xv -\xv^t),
\end{equation}
where $\Imat$ denotes the identity matrix and $\zeta$ must be
chosen to be equal to or greater than the maximum eigenvalue of ${\bf A}^{\top}{\bf A}$. For the Hadamard sensing matrix used in our camera, the maximum eigenvalue of ${\bf A}^{\top}{\bf A}$ is easily obtained. 
The update equation of $\xv^t$ in this Iterative Shrinkage Thresholding (IST) algorithm~\cite{Beck09IST} is given by:
\begin{equation}\label{eq:ISTxk}
\xv^{t+1} = \xv^t +  \frac{1}{\zeta} \Amat^{\top}(\yv - \Amat \xv^t).
\end{equation}

\subsection{Generalized Alternating Projection}
\label{Sec:GAP}
The Generalized Alternating Projection (GAP) algorithm proposed in~\cite{Liao14GAP}, which enjoys the anytime property and has been demonstrated high performance in video compressive sensing~\cite{Yuan14CVPR}, has the following update equation by using the Euclidean projection:
\begin{equation}\label{eq:GAPxk}
\xv^{t+1} = \xv^t +  {\bf A}^{\top}({\bf A A}^{\top})^{-1}(\yv - {\bf A} \xv^t).
\end{equation}
Under some condition of the sensing matrix ${\bf A}$, as the Hadamard matrix used in our system, $\bf A A^{\top}$ is the identity matrix and thus (\ref{eq:GAPxk}) is same as (\ref{eq:ISTxk}) with $\zeta =1$.

In addition to (\ref{eq:GAPxk}), aiming to speed-up the convergence, the authors in~\cite{Liao14GAP} have proposed the accelerated update equations
\begin{eqnarray}
\xv^{t+1} &=& \xv^t +  {\bf A}^{\top}({\bf A A}^{\top})^{-1}({\bf y}^t - {\bf A} \xv^t), \label{eq:GAPaccxk}\\
\yv^t &=& \yv^{k-1} + (\yv -\Amat \xv^{t-1}).
\end{eqnarray}
Better results have been achieved in our experiments using this accelerated GAP. 
\begin{center}
	\begin{algorithm}[htbp]
		\caption{LR-GMM-SLOPE}
		\begin{algorithmic}[1]
			\REQUIRE Measurements ${\yv}$, sensing matrix $\Amat$.
			\STATE Initial $\xv$.
			\FOR{$t=1$ \TO Max-Iter }
			\STATE Update $\xv$ by IST~\eqref{eq:ISTxk}, or GAP~\eqref{eq:GAPaccxk} or ADMM~\eqref{eq:x4_k+1}.
			\STATE Update related parameters in IST, GAP or ADMM.
			\STATE Learn a GMM (not low-rank) from $\xv$.
			\STATE Obtain the low-rank GMM via eigenvalue shrinkage thresholding~\eqref{eq:SigmaLowRank}.
			\STATE Update $\xv$ by the low-rank GMM using expectation in~\eqref{eq:txhatmean}.
			\ENDFOR
		\end{algorithmic}
		\label{algo:GMMslope}
	\end{algorithm}
\end{center}

\begin{center}
	\begin{algorithm}[htbp]
		\caption{LR-PLE-SLOPE}
		\begin{algorithmic}[1]
			\REQUIRE Measurements ${\yv}$, sensing matrix $\Amat$.
			\STATE Initial $\xv$.
			\FOR{$t=1$ \TO Max-Iter }
			\STATE Update $\xv$ by IST~\eqref{eq:ISTxk}, or GAP~\eqref{eq:GAPaccxk} or ADMM~\eqref{eq:x4_k+1}.
			\STATE Update related parameters in IST, GAP or ADMM.
			\STATE Update the Gaussian models (not low-rank) from $\xv$ via (\ref{eq:plemu_sig})-(\ref{eq:plesig}).
			\STATE Obtain the low-rank Gaussian models via \eqref{eq:SigmaLowRank}.
			\STATE Update $\xv$ by the low-rank Gaussian models using (\ref{eq:txv_k})-(\ref{eq:xv_i}).
			\ENDFOR
		\end{algorithmic}
		\label{algo:PLEslope}
	\end{algorithm}
\end{center}
\subsection{An ADMM Formulation}
\label{Sec:ADMM}
Under the GMM framework, we don't have the sparse variable $\zv$ as in (\ref{eq:P1}), the objective function can be formulated as:
\begin{equation}\label{eq:P3}
\xv = \arg\min_{\xv}\frac{1}{2}\|\yv - \Amat\xv\|^2_2 + \frac{\eta}{2}\sum_i \|{\bf R}_i \xv - \tilde{\xv}_i\|_F^2.
\end{equation}
where $\tilde{\xv}$ is obtained by the low-rank GMM model.
Following the procedure in (\ref{eq:P2}) by introducing the auxiliary variable \{$\wv, \vv$\}, we have
\begin{align}\label{eq:minxGMM}
(\xv,\wv,\vv)&=\underset{\xv,\wv,\vv}{\argmin}  \frac{1}{2} \|\yv - \Amat\xv\|^2_2  +  \frac{\eta}{2}\sum_i \|{\bf R}_i \wv - \tilde{\xv}_i\|_2^2  \nonumber\\
&~~+  \frac{\beta}{2}\|\xv-\wv + \vv\|_2^2 + {\rm const}
\end{align}
The optimization of (\ref{eq:minxGMM}) consists of the following iterations:
\begin{align}
\xv^{t+1} &:= \arg \min_{\xv} \frac{1}{2}\|\yv - \Amat\xv\|^2_2 +\frac{\beta}{2}\|\xv-\wv^t + \vv^t\|_2^2,
\label{eq:x4_k+1}\\
\wv^{t+1}&:= \frac{\eta}{2}\sum_i \|{\bf R}_i \wv - \tilde{\xv}^t_i\|_2^2 + \frac{\beta}{2}\|\xv^{t+1}-\wv + \vv^t\|_2^2, \label{eq:w4_k+1}\\
\boldsymbol{v}^{t+1}&:= \boldsymbol{\vv}^{t} + (\xv^{t+1} - \wv^{t+1}).
\label{eq:v4_k+1}
\end{align}
where the update of $\tilde{\xv}$ is given by the low-rank GMM in (\ref{eq:xv_pdf})-(\ref{eq:txhatmean}).
Under the sensing matrix considered here in our work, $\Amat\Amat^{\top} = \Imat$, 
the solution of (\ref{eq:x4_k+1}) is given by (\ref{eq:xp3_k+1}).
Eq. (\ref{eq:w4_k+1}) can be solved by
\begin{align}\label{eq:wp4_k+1}
\wv^{t+1} = (\eta\sum_i{\bf R}_i^{\top} {\bf R}_i + \beta \Imat)^{-1}[\beta(\xv^{t+1}+ \vv^t) + \eta\sum_i{\bf R}^{\top}_i \tilde{\xv}_i^t]
\end{align}
Similar to (\ref{eq:wnk+1}), $\wv^{t+1}$ can be solved element-wise but in one shot.

The proposed low-rank GMM algorithm, integrated with the three approaches to update $\xv$ (projecting $\yv$ to $\xv$), constitutes the LR-GMM-SLOPE algorithm summarized in Algorithm~\ref{algo:GMMslope}.
Similarly, when the low-rank constraint is imposed on the PLE, we obtain the LR-PLE-SLOPE in Algorithm~\ref{algo:PLEslope}.

\begin{table*}[htbp!]
	\caption{Reconstruction PSNR ({dB}) of different images with diverse algorithms at various CSr.}
	\centering{\scriptsize 
		\begin{tabular}{c|c|cccccccc}
			\hline Image & Method
			&  CSr$= 0.03$  & CSr$= 0.04$ & CSr$= 0.05$ & CSr$= 0.06$ & CSr$= 0.07$ & CSr$= 0.08$ & CSr$= 0.09$ & CSr$= 0.1$ \\
			\hline \hline
			& Proposed & {\bf 21.7416} & {\bf 22.7603} & {\bf 23.5292} & {\bf 24.1613} & {\bf 24.6844} & 25.2192 & {\bf 25.6663} & {\bf 26.0390}  \\
			& NLR-CS & 20.4887 & 21.8187 & 20.9706 & 21.6355  & 24.5237 & {\bf 25.4407} & 25.5486  & 25.7166  \\
			& D-AMP & 19.5424  & 20.4412 & 21.6364 & 22.2731   & 23.0846 & 23.8539 & 24.5367 & 25.0486   \\
			& GAP-w & 20.0870 & 20.9902 & 21.6160  & 22.2509  & 22.6795 & 23.0811 & 23.4164  & 23.7371  \\
			barbara	& TVAL3& 18.1065 & 19.3655 & 20.6883  & 21.4713  & 22.0738 & 22.8414 & 23.0771  & 23.4208   \\
			\hline
			& Proposed & {\bf 22.7729} & {\bf 22.8600} & {\bf 24.6319} & {\bf 25.2106} & {\bf 25.9662} & {\bf 26.5287} & {\bf 27.1382} & {\bf 27.7893}  \\
			& NLR-CS & 21.8342 & 21.4521 & 24.3866 & 16.9015  & 22.2784 & 24.5024 & 23.7634 & 23.7528   \\
			& D-AMP & 19.7960  & 20.9017 & 21.9684 & 22.8501   & 23.5793 & 24.2540 & 24.9395 & 25.5490\\
			& GAP-w & 20.6312 & 21.2215 & 21.8709  & 22.3753  & 22.8921 & 23.3859 & 23.7879  & 24.1355   \\
			boat	& TVAL3 & 18.2669 & 19.4750 & 20.2451  & 21.5041  & 21.9831 & 22.3542 & 22.7917  & 23.1918   \\
			\hline
			& Proposed & {\bf 21.3706} & {\bf 22.1542} & {\bf 22.8186} & {\bf 23.4861} & {\bf 24.0118} & {\bf 24.4425} & {\bf 24.8875} & {\bf 25.1473}  \\
			& NLR-CS & 12.0313 & 16.6251 & 16.0262 & 17.7892  & 17.8507 & 18.7355 & 19.2338 & 21.5345   \\
			& D-AMP & 18.3514  & 19.3660 & 20.3618 & 21.2030   & 22.0050 & 22.7786 & 23.3635 & 24.0456  \\
			& GAP-w & 19.1932 & 19.7725 & 20.2935  & 21.1823  & 21.5875 & 21.8644 & 22.1527  & 23.8935  \\
			cameraman	& TVAL3 & 17.8787 & 18.6441 & 20.1835 & 20.2041 & 20.8832  & 21.2041 & 21.2917& 21.4903     \\
			\hline
			& Proposed & 28.8157 & 29.8322 & 30.5442 & 31.5101 & 32.0363 & 32.5592 & 33.2439 & 33.5244 \\
			& NLR-CS & {\bf 29.7873} & {\bf 30.4427} & {\bf 31.9318} &  {\bf 33.4732}  & {\bf 34.0312} & {\bf 34.8470} & {\bf 34.9993} & {\bf 34.4573}   \\
			& D-AMP & 22.3218  & 24.4925 & 26.1075 & 27.5484   & 28.9518 & 30.2165 & 31.4011 & 32.3636 \\
			& GAP-w & 24.5972 & 25.5398& 26.2436 & 26.8443  & 27.3006& 27.8175 & 28.2215  & 26.6864   \\
			foreman	& TVAL3 & 19.3428	& 20.9050	& 23.1337 &	24.1152&	24.9492&	25.42780 &	25.9380&	26.5072 \\
			\hline
			& Proposed & {\bf 26.7712} & 	28.5080 &	29.5543 &		{\bf 30.1185}&		{\bf 31.0571}&	31.5114 &		{\bf 32.1238} &		{\bf 32.6598} 
			\\
			& NLR-CS & 25.8497 &	{\bf 28.1219} &	{\bf 30.6011}  &	27.6256 &	30.1692 &		{\bf 31.5297}  &	28.8471 &	29.2505  \\
			& D-AMP & 21.7965 &	23.7841 &	25.2089 &	26.6130& 	27.8643 &	28.9586 &	30.0785 &	31.1040 \\
			& GAP-w & 23.0012 & 	23.8285 &	24.5941 &	25.3161 &	25.8258 &	26.3531&	26.8563&	27.2498 \\
			house	& TVAL3 & 19.0674 &	20.6872 &	21.7140 &	23.4273&	23.7382&	24.0901 &	24.5912 &	25.1538\\
			\hline
			& Proposed & 	{\bf 22.9046} &		{\bf 23.9496} &	24.6221 &		{\bf 25.3661} &		{\bf 25.9910} &		{\bf 26.3633} &		{\bf 26.9981} &		{\bf 27.3694}
			\\
			& NLR-CS & 22.6851 &	22.6874 &		{\bf 24.8531} &	23.2517 &	21.2840 &	24.2211 & 25.7213 &	27.2040 \\
			& D-AMP & 20.4629 &	21.7906 &	22.5929 &	23.6507 &	24.2494 &	24.7895&	25.3561 &	25.8032  \\
			& GAP-w & 20.7381 &	21.4840 &	22.2093 &	22.6754 &	23.0767  &	23.5415 &	23.9218 &	24.2559 \\
			lena	& TVAL3 & 19.1813 &	19.7752 &	20.7421 &	21.7209 &	22.1211 &	22.6207 &	23.0676 &	23.6665  \\
			\hline
			& Proposed & 	{\bf 19.2072} &		{\bf 20.1870} &		{\bf 21.3428} &		{\bf 22.3022} &		{\bf 22.9918} &		{\bf 23.6285} &		{\bf 24.2536} &		{\bf 24.7206} 
			\\
			& NLR-CS & 16.1196 &	17.5697 &	17.1601 &	14.6306 &	15.5702 &	18.8192 &	20.8003 &	20.8957  \\
			& D-AMP & 16.8849 &	17.6731 & 18.8636 &	19.7654 &	20.8845&	21.8400&	22.7742 &	23.4913 \\
			& GAP-w & 17.3904 &	18.0572 &	18.8062 &	19.3374 &	19.8313 &	20.2766 &	20.7840 &	21.1890  \\
			monarch	& TVAL3 & 16.2031 &	17.4510 &	18.0521 &	18.6358 &	19.0864 &	19.6194 &	19.9829&	20.4556 \\
			\hline
			& Proposed & {\bf 23.1402} &	{\bf 24.1143} &	{\bf 24.9261} &	{\bf 25.5033} &	{\bf 26.4840} &	{\bf 26.9776} &	{\bf 27.4935} & 	{\bf 28.1257} 
			\\
			& NLR-CS & 20.4401 &	20.9168 &	22.6560&	22.1715&	22.1582 &	22.4706&	24.1134&	26.3153\\
			& D-AMP & 19.5304 &	20.8630 &	21.9746 & 	22.8995 &	23.9191&	24.6991 &	25.5513 &	26.5082 \\
			& GAP-w & 20.9576 &	21.9057 &	22.5781&	23.1853 &	23.7812 &	24.2326 &	24.7597 &	25.1206 \\
			parrot	& TVAL3 & 18.2386 &	19.7337&	21.4420&	22.1800&	21.8906&	22.6345&	22.9478&	23.5418  \\
			\hline
			& Proposed & {\bf 23.3141} &	{\bf 24.4109} &	{\bf 25.2446} &	{\bf 25.9507} &	{\bf 26.6528} &	{\bf 27.1457} &	{\bf 27.6911 }&	{\bf 28.1719} 
			\\
			& NLR-CS & 21.1545 &	22.4543 &	23.5732 &	22.1848 &	23.4832 &	25.0708 &	25.2534&	26.1408 \\
			& D-AMP & 19.8358 &	21.1640 &	22.3393 &	23.3504 &	24.3173 &	25.1738 &	26.0001 &	26.7392 \\
			& GAP-w & 20.8245 &	21.5999 &	22.2766 &	22.8427 &	23.3212 &	23.7845 &	24.2015 &	24.5660  \\
			average	& TVAL3 & 18.2857 &	19.5046 &	20.7751 &	21.6573 &	22.0907 &	22.5990&	22.9610 &	23.4285 \\
			\hline
		\end{tabular}}
		\label{Table:sim_PSNR}
	\end{table*}

	\begin{table*}[htbp!]
		\caption{Reconstruction PSNR ({dB}) of RGB images with diverse algorithms at various CSr.}
		\centering{\scriptsize 
			\begin{tabular}{c|c|cccccccc}
				\hline Image & Method
				&  CSr$= 0.03$  & CSr$= 0.04$ & CSr$= 0.05$ & CSr$= 0.06$ & CSr$= 0.07$ & CSr$= 0.08$ & CSr$= 0.09$ & CSr$= 0.1$ \\
				\hline \hline
				\multirow{5}{*}{\includegraphics[width=1.3cm, height=1.3cm]{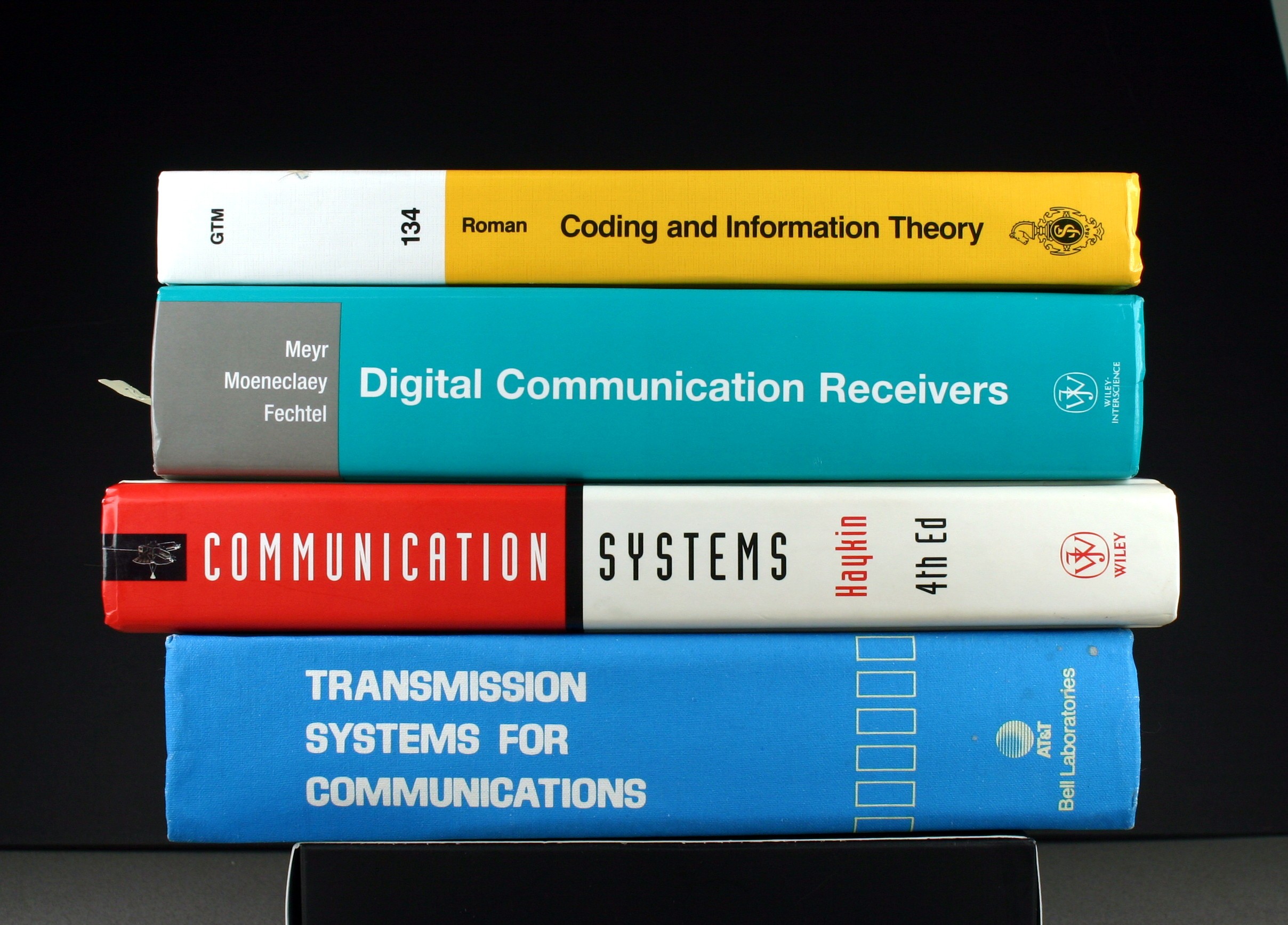}} & Proposed & {\bf 22.0258} &	{\bf 23.0449}&	{\bf 23.8714}&	{\bf 24.4931} &	{\bf 25.0737}&	{\bf 25.6280}&	{\bf 26.0476}&	{\bf 26.4061} \\
				& NLR-CS & 12.9754 &	13.8812 &	19.0401&	17.2073 &	17.5413&	18.9149 &	22.0660&	21.8692    \\
				& D-AMP & 17.3405 &	18.9956 &	20.6016 &	21.8303&	22.8833&	23.8037&	24.6248&	25.1828  \\
				& GAP-w & 18.9621 &	19.8426 &	20.5282 &	21.0880&	21.5961&	21.9764&	22.2893&	22.5441 \\
				& TVAL3& 15.9549 &	17.0403&	18.0474&	18.9911 &	19.4949 &	20.1248&	20.4482&	20.8150   \\
				\hline
				\multirow{5}{*}{\includegraphics[width=1.3cm, height=1.3cm]{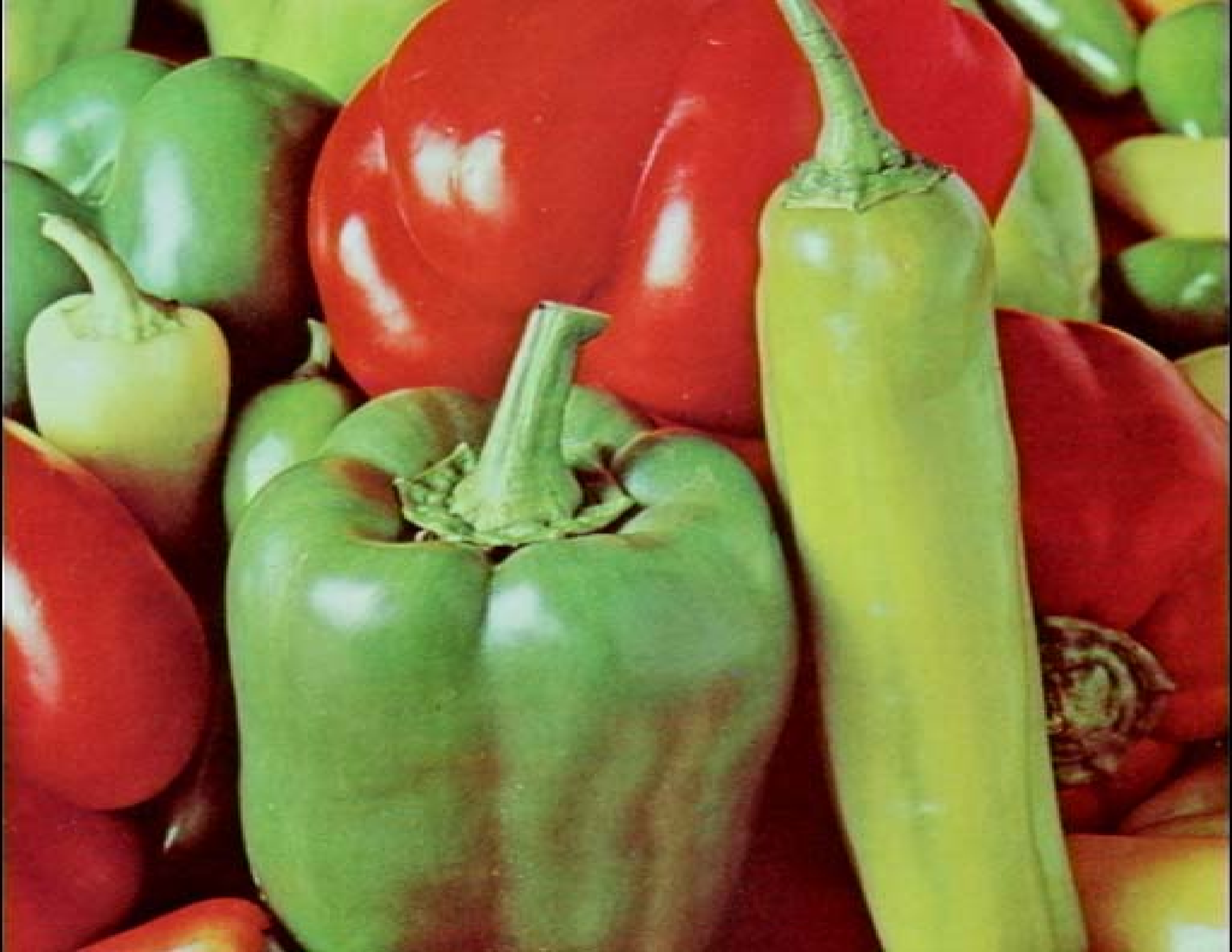}}& Proposed & {\bf 22.4159}&	{\bf 23.4245}&	{\bf 24.4111}&	{\bf 25.3072} &	{\bf 26.1026}&	{\bf 26.7247}&	{\bf 27.3431}&	{\bf 27.8475} \\
				& NLR-CS & 15.5817 & 19.0300 &	19.9201 &	19.4926 &	21.1154 &	23.7973&	24.8067&	23.6006\\
				& D-AMP & 17.3405 &	18.9956 &	20.6018 &	21.8303&	22.8833&	23.8037&	24.6248 & 25.1828 \\
				& GAP-w & 18.9621 &	19.8426 &	20.5282 &	21.0880 &	21.5961&	21.9764&	22.2893&	22.5441  \\
				& TVAL3 & 15.9549 &	17.0403&	18.0474&	18.9911 &	19.4949 &	20.1248&	20.4482&	20.8150  \\
				\hline
				
				\hline
			\end{tabular}}
			\label{Table:sim_PSNR_rgb}
		\end{table*}
\section{Simulation Results}
\label{Sec:simResults}
We test the proposed algorithm on simulation datasets with 2D images.
The proposed algorithm is compared with other leading algorithms
1) TVAL3~\cite{Li13COA}, 2) GAP based on wavelet~\cite{Liao14GAP}, 3) DAMP~\cite{Mertzler14Denoising} with BM3D denoising, and 4) NLR-CS~\cite{Dong14TIP}, which explores the low rank of similar patches.
State-of-the-art results have been obtained by~\cite{Mertzler14Denoising,Dong14TIP}.
The Gaussian components in our mixture model is set to $K=6$ for all the experiments and the analysis of this number is provided in Section~\ref{Sec:GMM_K}. When updating $\xv$, accelerated GAP in (\ref{eq:GAPaccxk}) is used and the comparison of different approaches is shown in Section~\ref{Sec:compx}.
We obtained the low-rank GMM by setting the rank of each Gaussian component learned via the EM algorithm to the half of the full rank $\gamma_k = 0.5P$, where P = 64 for the patch size $8\times 8$ used in this paper.
The proposed algorithm is further compared with JPEG compression in Section~\ref{Sec:JPEG}.

\begin{figure}[htbp]
	\centering
	\includegraphics[width=0.11\textwidth]{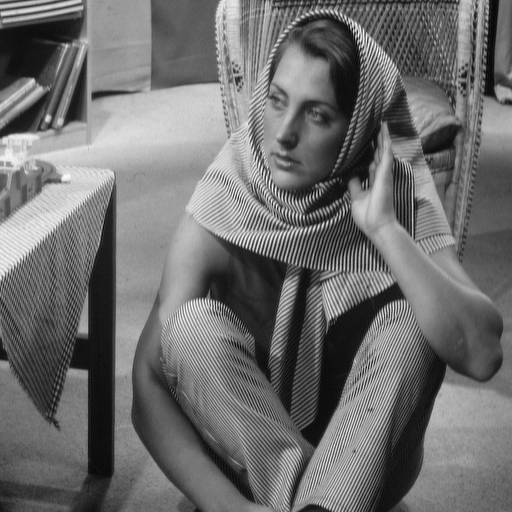}
	\includegraphics[width=0.11\textwidth]{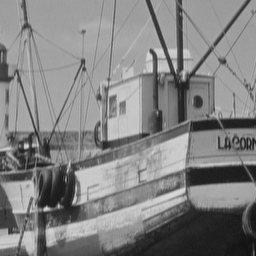}
	\includegraphics[width=0.11\textwidth]{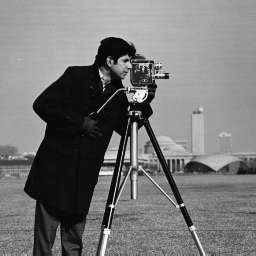}
	\includegraphics[width=0.11\textwidth]{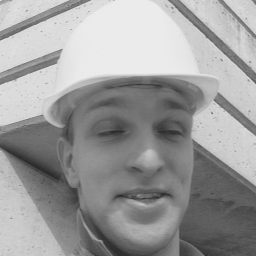}\\ 
	\vspace{1mm}
	\includegraphics[width=0.11\textwidth]{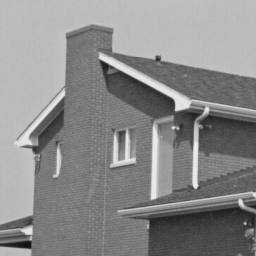}
	\includegraphics[width=0.11\textwidth]{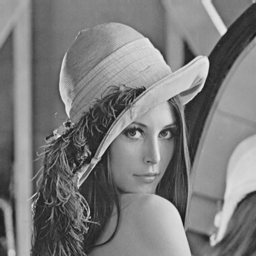}
	\includegraphics[width=0.11\textwidth]{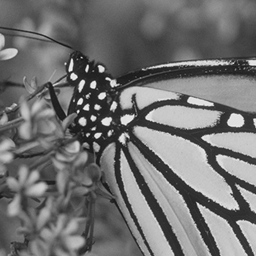}
	\includegraphics[width=0.11\textwidth]{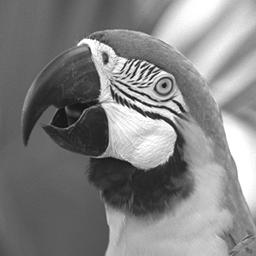}
	\caption{Images used for CS experiment, \{barbara, boat, cameraman, foreman, house, lena, monarch, parrot\}.}
	\label{fig:CSimage}
\end{figure}

\begin{figure}[htbp!]
	\centering
	\includegraphics[width=0.48\textwidth]{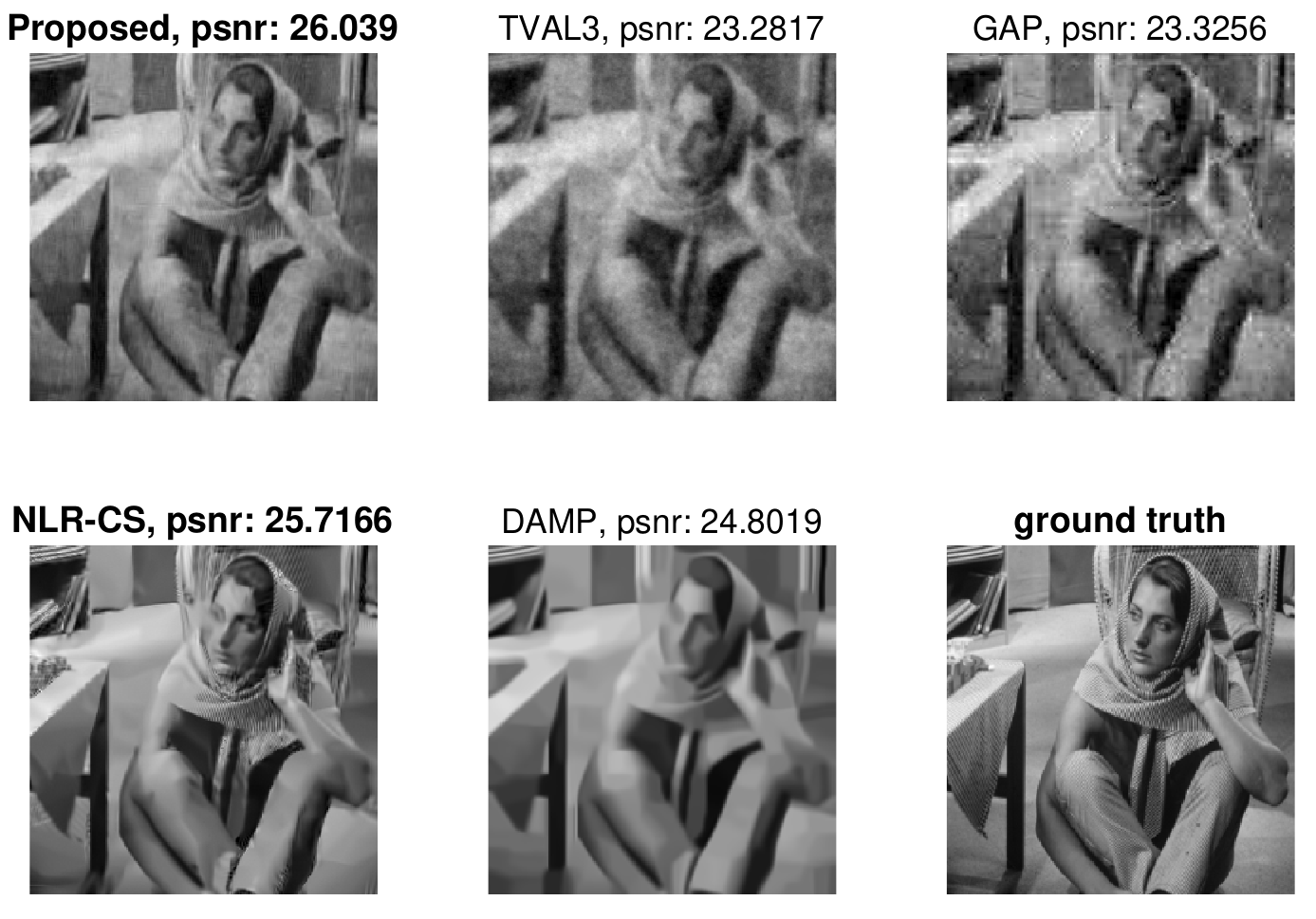}
	\vspace{-3mm}
	\caption{Reconstruction results of different algorithms at CSr$=0.1$, image size $256\times 256$.}
	\label{fig:baba01}
\end{figure}
%
Following the formulation of the lensless camera~\cite{Huang13ICIP}, the permuted Hadamard matrix is used as the sensing matrix.  
Each image is resized to $256\times 256$ and these images are shown in Figure~\ref{fig:CSimage}. 
The CSr is defined as:
\begin{equation}
\text{CSr} = \frac{{\text {number of rows in }} \Amat}{{\text {number of columns in }} \Amat}
\end{equation}
where the number of columns in $\Amat$ is equivalent to the total pixel number of the image.
The sensing matrix is constructed from rows of a Hadamard matrix of order $N=2^{16}$. The columns of the Hadamard matrix is permuted according to a predetermined random permutation (the same permutation is used in the real data captured by our lensless camera).
For each CSr, we use the first CSr$\times N$ rows of the column-permuted Hadamard matrix as the sensing matrix.
By selecting some other rows of the permutated Hadamard matrix, we can get better results~\cite{Romberg08SPM}. However, here we just select the top rows to be consistent to the implementation of our lensless camera.
Note that in this case, $\Amat\Amat\ts$ is an identity matrix and it is very fast to use accelerated GAP update for $\xv$ in~\eqref{eq:GAPaccxk}. We observed that best results are obtained by the LR-GMM-SLOPE with GAP updates of $\xv$ and these results are reported in this section.
For the comparison of IST, GAP and ADMM, please refer to Section~\ref{Sec:compx}.

Since when CSr=0.1, good results have been achieved for most images (see Figure~\ref{fig:baba01} for one example), we here spend more efforts on the extremely low CSr, in particular CSr$<0.1$,  which may be of interest to the real applications that are used to detect anomalous events~\cite{Jiang12Inverse} without caring too much about the image quality.
The results are summarized in Table~\ref{Table:sim_PSNR}. We can observe that best results are obtained by the proposed aglorithm, NLR-CS or D-AMP. When CSr is less than 0.1, the proposed algorithm usually provides best results except ``foreman", where NLR-CS is the best.
We also observed that NLR-CS is very sensitive to the parameters and sometimes the PSNRs are not linearly increasing as the CSr increases, while the other algorithms including the proposed do not have this problem.
On average, our proposed algorithm works best when CSr$\le 0.1$.
Though did not reported here, when CSr$>0.1$, the proposed algorithm also provides comparable or better results than D-AMP and NLR-CS. But we need to tune the noise parameter used in $\nv$, while for all the results presented here, we set it to the same value $\Emat = 10^{-5} \Imat$.
In addition, we need to tune the rank thresholds $\gamma_k$, for which we have observed that a higher CSr requires a larger $\gamma_k$.

\begin{figure}[htbp!]
	\centering
	\includegraphics[width=0.48\textwidth]{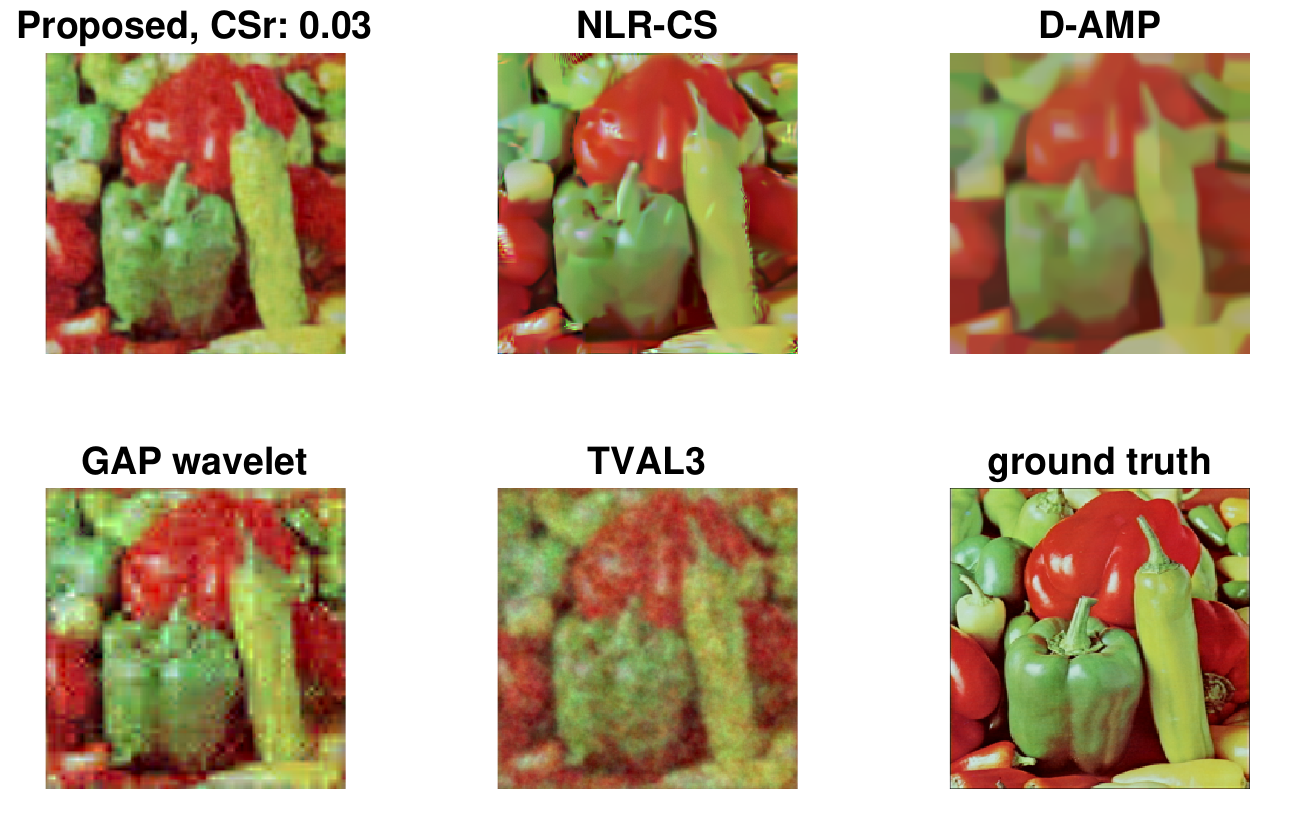}
	\vspace{-3mm}
	\caption{Reconstruction results of different algorithms at CSr$=0.03$.}
	\label{fig:peppers003}
\end{figure}
In addition to the grayscale images tested above as in other papers, we also conduct our proposed algorithm on the RGB images with results shown in Table~\ref{Table:sim_PSNR_rgb}.
Three sensors are simulated to capture the R, G and B components of the image. 
The ``book" scene corresponds to the real data captured by our lensless camera.
Again, our proposed algorithm provides the best results. One example with CSr$=0.03$ is shown in Figure~\ref{fig:peppers003}. It can be seen that our algorithms provide more details than NLR-CS and D-AMP; both of them presents ``blob" artifacts.

\subsection{Compare to JPEG Compression}
\label{Sec:JPEG}
\begin{table}[tbp!]
	\caption{JPEG compression at different qualities compared with the proposed compressive sensing recovery}
	\centering
	\begin{tabular}{|c|c|c||c|c||c|}
		\hline \multicolumn{3}{|c||}{JPEG Compression} & \multicolumn{2}{|c||}{CS Reconstruction} & Difference  \\
		\hline   & Size  & PSNR 
		&   &  PSNR  & PSNR \\
		Quality & (bytes) & (dB) & CSr  & (dB) & (dB)\\
		\hline \hline
		1 & 1,563 & 22.0683 & 0.0334 & 22.1507 & {\bf -0.0823}\\
		\hline
		3 & 1,716 & 22.7278 & 0.0367 & 22.4125 & 0.3154\\
		\hline
		5 & 2,153 & 24.8416 & 0.0460 & 23.2433 & 1.5983\\
		\hline
		7 & 2,559 & 26.0924 & 0.0547 & 23.8426 & 2.2548 \\
		\hline
		9 & 2,946 & 26.9623 & 0.0630 & 24.3308 & 2.6315\\
		\hline
		10 & 3,141 & 27.2853 & 0.0672 & 24.5945 & 2.6908\\
		\hline
		12 & 3,494 & 27.8379 & 0.0747 & 24.9514 & 2.8866 \\
		\hline
		14 & 3,815 & 28.2981 & 0.0816 & 25.3233  & 2.9748 \\
		\hline
		16 & 4,160 & 28.6912 & 0.0890 & 25.6880  & 3.0032\\
		\hline
		18 & 4,478 & 29.0306 & 0.0958 & 25.9230 & 3.1077\\
		\hline
		20 & 4,752 & 29.3411 & 0.1016 & 26.1267 & 3.2144\\
		\hline	
	\end{tabular}
	\label{Table:JPEG_PSNR}
\end{table}
\begin{figure}[htbp!]
	\centering
	\includegraphics[width=0.49\textwidth]{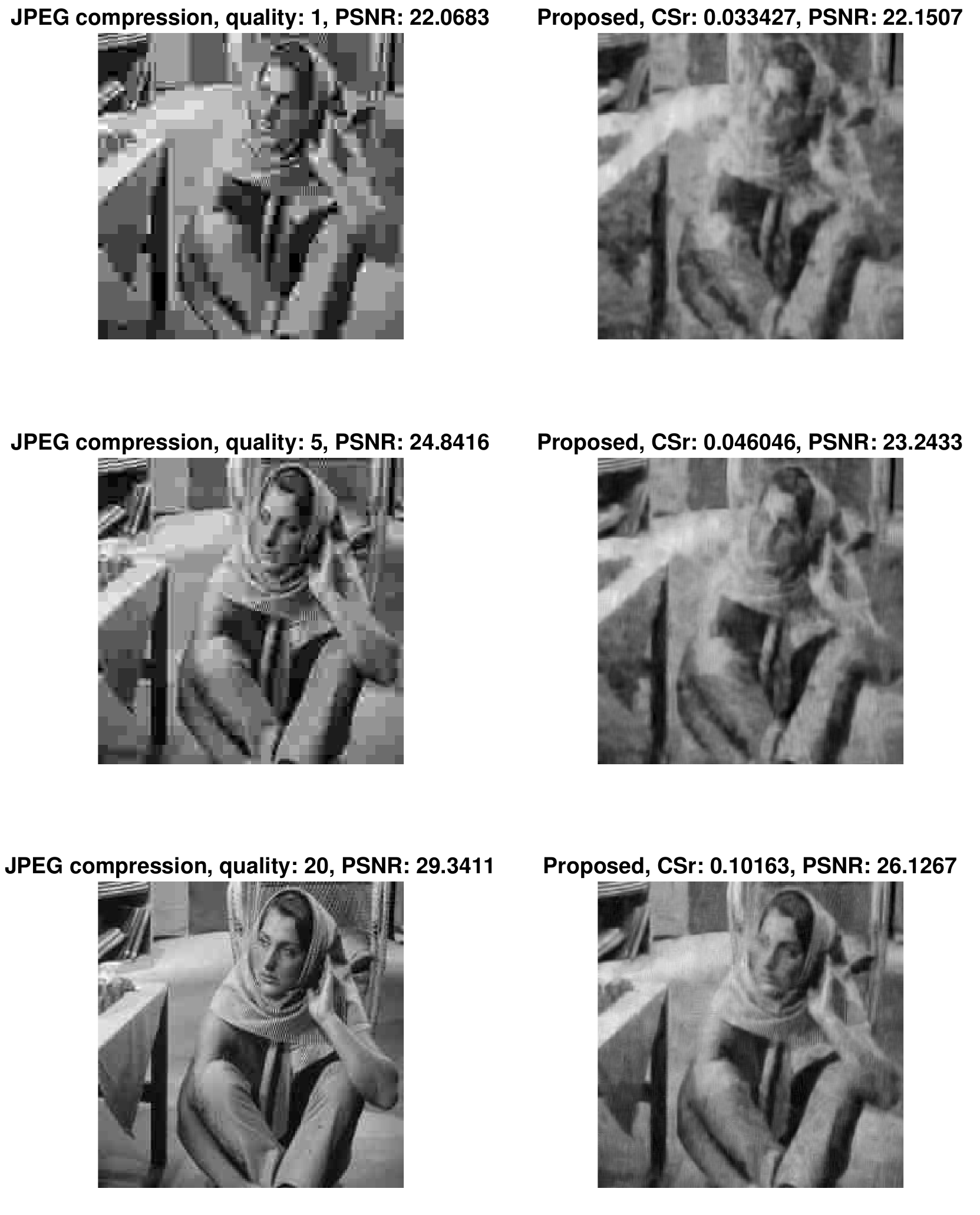}\\
	\vspace{-3mm}
	\caption{Example images: JPEG compared with the proposed algorithm. }
	\label{fig:JPEG_comp}
\end{figure}
We now compare LR-GMM-SLOPE under the compressive sensing framework with the JPEG compression, which is based on the sparsity of the DCT coefficients of $8\times 8$ blocks (non-overlapping patches).
We first use an PNG file as the ground truth and then use the script within MATLAB ``imwrite($\cdot$)" by choosing 8-bits `jpeg' compression with different qualities (100 denotes the highest quality).
We treat the quality 100 as the standard full file size. For the `Barbara' image we used here, PSNR = 58.47dB (w.r.t. the PNG file) and the file size is 45.6KB at quality 100.
The compressed image is obtained by changing the compression quality from 1 to 100 and we compare the file size with the full size at quality 100, computing the CSr used in this paper.

Table~\ref{Table:JPEG_PSNR} summarizes the results of JPEG compression compared with the results obtained by our algorithm.
This is a rough, high level comparison because JPEG also performs an entropy encoding after the DCT transform and quantization, while in our method, the number of compressive measurements is compared with the number of total pixels, and we did not consider the entropy coding on quantized measurements.
We intend to take the effect of the entropy encoding in JPEG out of the comparison by computing the JPEG compression ratio as compared to the quality 100. 
When the compression is high (lower CSr), the gap between our approach and the JPEG compression is small. 
It is worth noting that when CSr=0.0334, our algorithm performs better than JPEG.
When the compression gets lower, the gap becomes larger.
One possible reason is that when JPEG is performed on the image, the ground truth is available and it is very easy to capture useful information from the truth.
However, under the compressive sensing framework and using the current algorithm, increasing a few number of measurements can help the reconstruction, but not that significantly. 
Example images can be found in Figure~\ref{fig:JPEG_comp}. It can be seen that JPEG compression has obvious block artifacts while the results of the proposed algorithm become better progressively with increasing number of measurements.

Furthermore, in JPEG compression, if we lose some bits, we may not be able to decode entire blocks. By contrast, in our compressive sensing framework, if we lose some measurements, we can still reconstruct the image, maybe not at a high fidelity.

\begin{figure}[htbp!]
	\centering
	\includegraphics[width=0.48\textwidth]{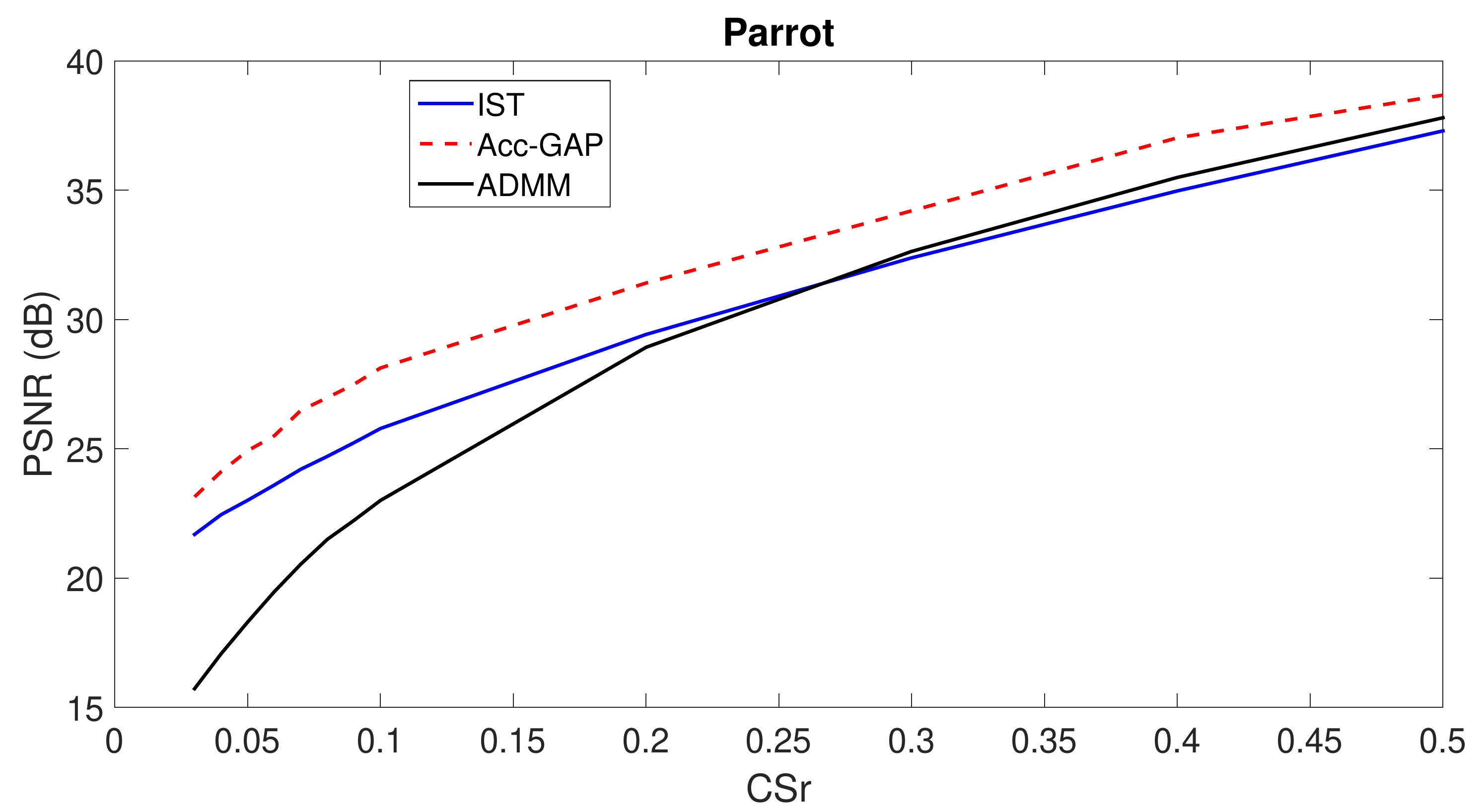}
	\vspace{-3mm}
	\caption{One example (parrot) to compare different updating rules (IST, Acc-GAP, ADMM) of $\xv$.}
	\label{fig:comp_admm_gap}
\end{figure}
\subsection{Comparison of Different Update Rules for $\xv$}
\label{Sec:compx}
We provide three approaches in Section~\ref{Sec:IST} to update $\xv$ in order to minimize the objective function in (\ref{eq:Jx}). Now we compare these three updates via experiments.
We emphasize again that we are using the permuted Hadamard matrix as the sensing matrix and thus $\Amat\Amat\ts$ is an identity matrix. Therefore, updating $\xv$ via GAP in (\ref{eq:GAPxk}) is same as updating $\xv$ via IST in (\ref{eq:ISTxk}). However, the accelerated GAP in (\ref{eq:GAPaccxk}) provides best results in our experiments.
Without tunning the ADMM parameters carefully, we compare these three update methods with different images at various CSr, and one example is shown in Figure~\ref{fig:comp_admm_gap}.
It can be observed that the accelerated GAP update always provides the best result and when CSr is low, IST is better than ADMM. When CSr is getting larger, ADMM becomes better than IST.
Because of this, all the results reported in this paper is generated by the accelerated GAP update.

\begin{figure}[htbp!]
	\centering
	\includegraphics[width=0.48\textwidth]{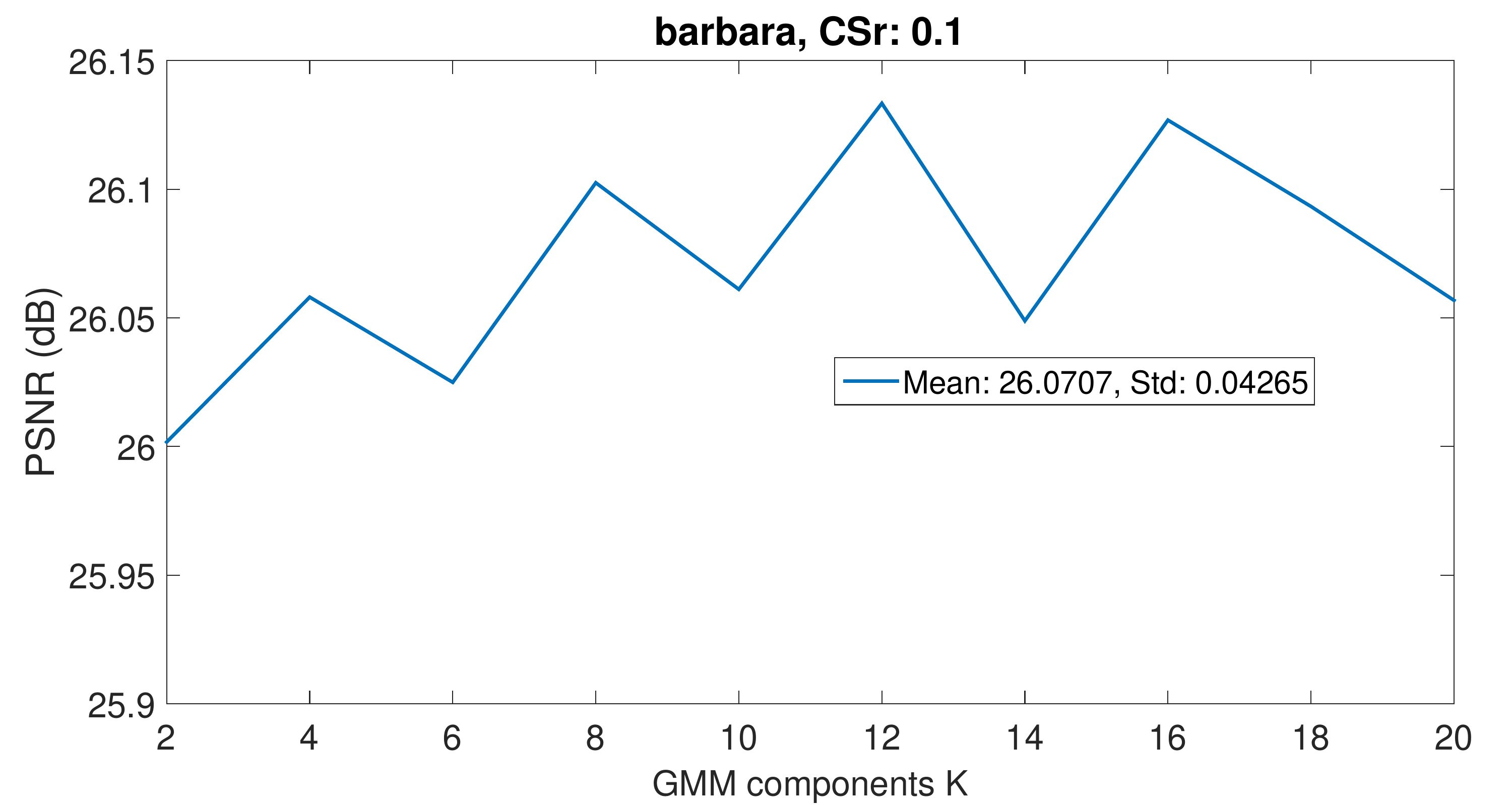}
	\vspace{-3mm}
	\caption{Reconstruction PSNR with different number of GMM components ($K$). Barbara is used with CSr = 0.1.}
	\label{fig:GMM_K}
\end{figure}
\subsection{Different Number of Gaussian Mixture Components}
\label{Sec:GMM_K}
One problem of using GMM is how to set the component number $K$. As we are using the mixture model, each patch is represented by the posterior distribution, another GMM.
Therefore, selecting this $K$ is not as critical as in the PLE~\cite{Yu12IPT}.
An alternative way to infer this $K$ is utilizing the manifold factor analysis model as developed in~\cite{Chen10SPT}.
Hereby, we investigate this point empirically by using the ``barbaba" image as used before with different number of $K\in [2, 20]$.
The results at CSr$=0.1$ are shown in Figure~\ref{fig:GMM_K}.
It can be seen that our algorithm is not sensitive to this $K$, since the standard deviation of the PSNRs with different $K$ is only 0.04265dB compared with the mean value 26.0707dB.

\begin{figure}[htbp!]
	\centering
	\includegraphics[width=0.48\textwidth]{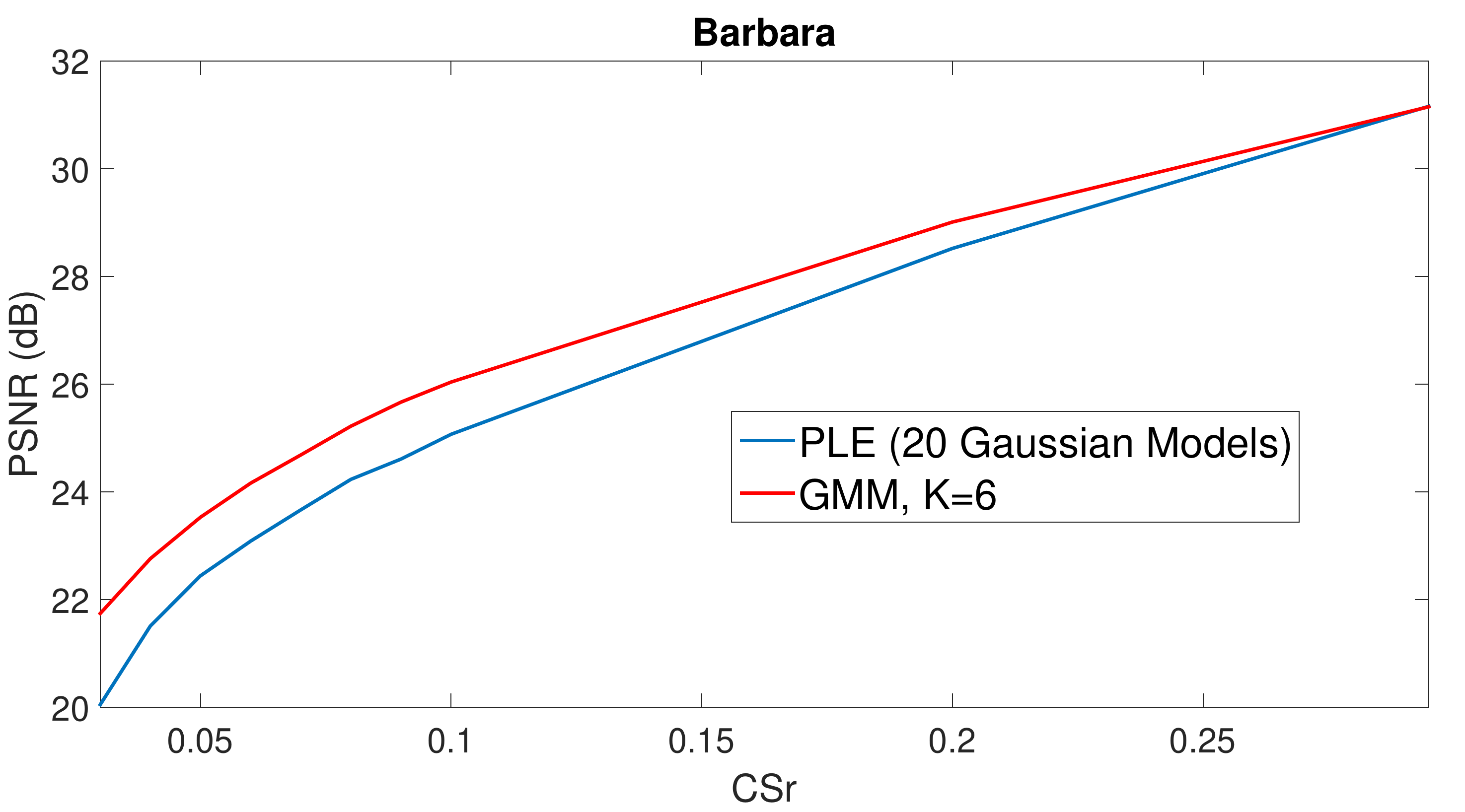}
	\vspace{-3mm}
	\caption{One example (barbara) to compare LR-PLE-SLOPE (20 Gaussian models) with LR-GMM-SLOPE ($K=6$). }
	\label{fig:GMM_PLE}
\end{figure}
\subsection{GMM vs. PLE and Computational Time}
\label{Sec:GMM_PLE}
As mentioned earlier, when we consider that each patch is drawn from a single Gaussian component, the proposed approach degrades to the PLE.
We verify the performance of LR-PLE-SLOPE compared with the LR-GMM-SLOPE in Figure~\ref{fig:GMM_PLE}. It can be seen that the GMM always performs better than the PLE at lower CSr. When CSr is getting larger, they start to perform similarly.

Regrading the computational time, our algorithm is similar to NLR-CS. If a warm start is used to initialize the $\hat{\xv}$, we can obtain a good reconstruction within 20 iterations.
One $256\times 256$ grayscale image reconstruction at CSr$=0.1$ takes around 1 minute on an i7CPU with 24G RAM. Similar time is required for the LR-PLE-SLOPE but it needs more memory. 
While the most time consumption of LR-GMM-SLOPE is the EM training of GMM, the LR-PLE-SLOPE requires a long time for the model selection, and it usually needs more Gaussian components (\ie, 20) than the GMM to get good results.

\section{Real Data Results for the Lensless Camera}
\label{Sec:realResults}
We now verify our proposed algorithm on the real data captured by our lensless camera~\cite{Huang13ICIP}, which is composed of an aperture assembly and a single sensor (a photodiode) to capture grayscale images; it can also be a RGB sensor to capture color images. 
The aperture assembly implements the sensing matrix and we programmed it to be the permuted Hadamard matrix. By capturing the scene with different sensing matrices, we obtain the measurement vector $\yv$.
We implemented the aperture assembly with a transparent LCD and thus we can control the image resolution by merging the neighboring pixels.  
\begin{figure}[htbp!]
	\centering
	\includegraphics[width=0.48\textwidth]{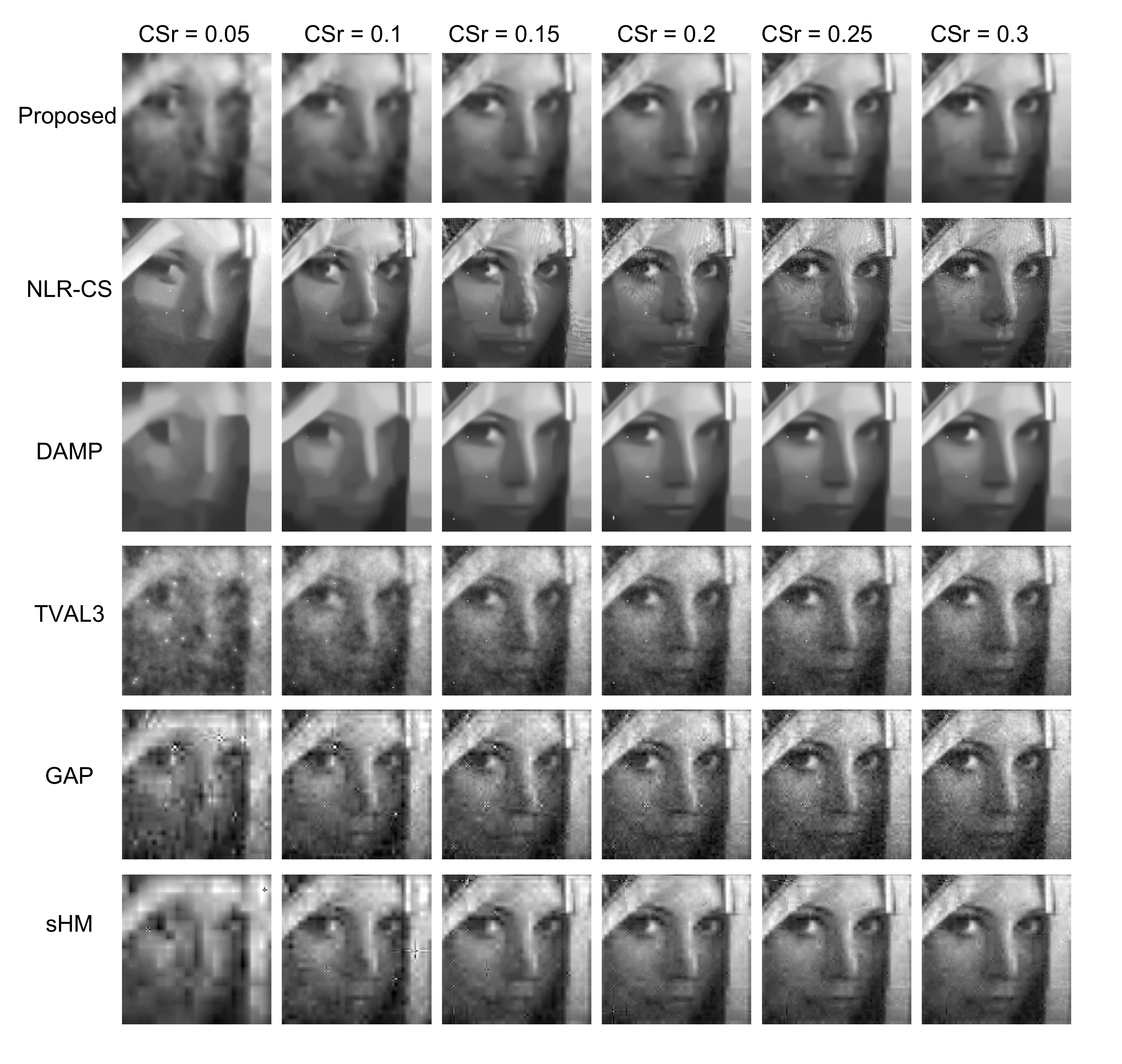}\\
	\hspace{0.35mm}\includegraphics[width=0.48\textwidth]{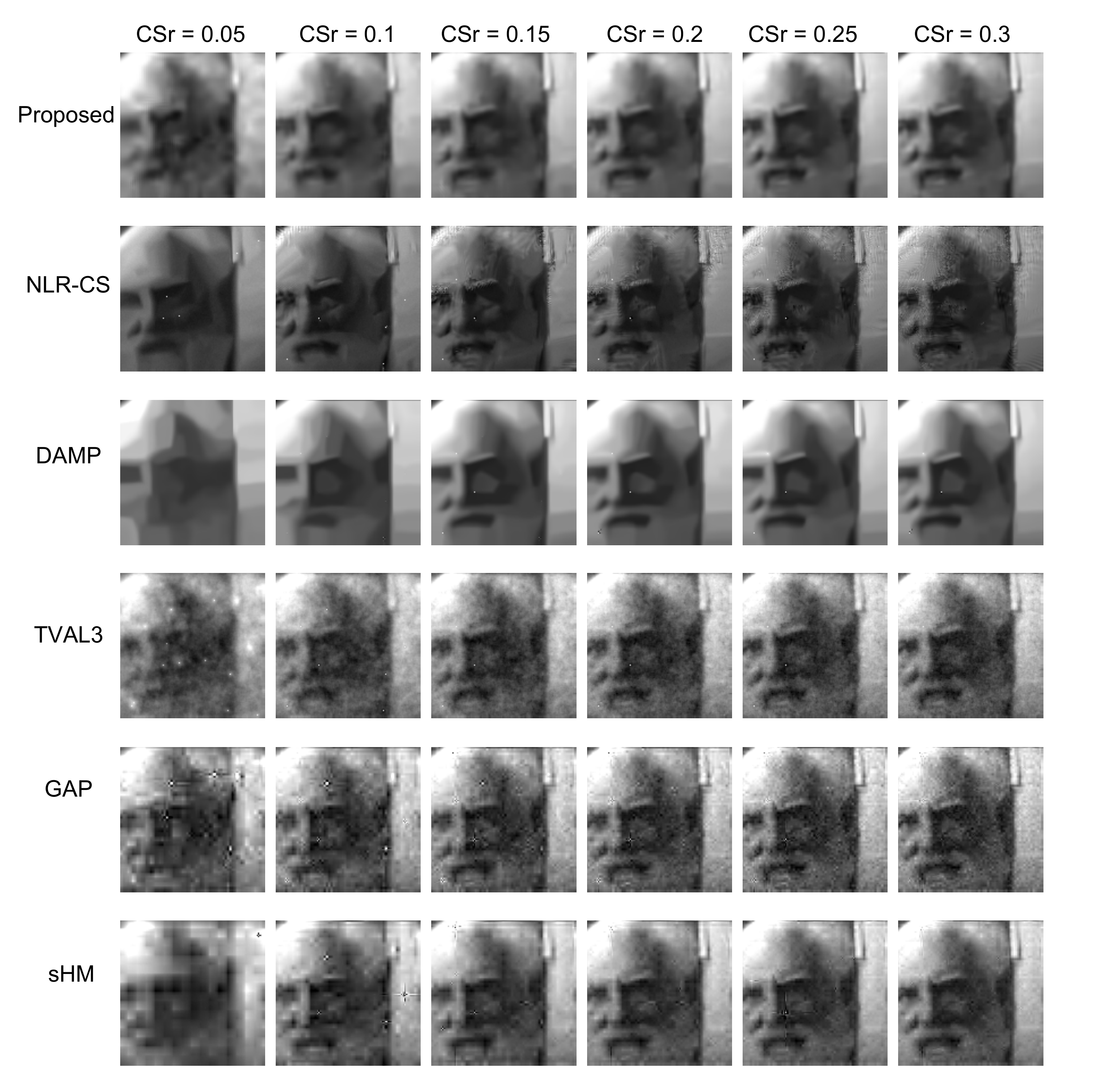}
	\vspace{-2mm}
	\caption{Real data: reconstruction results at different CSr with the diverse algorithms. The image is of size $128\times 128$. Two photos (top: Lena, bottom: Alexander Graham Bell) are used as the scene.}
	\label{fig:Lena_real}
\end{figure}
\subsection{Gray-Scale Images}
We first consider the case with gray scale sensor and the image resolution of $128\times 128$. 
To capture compressive measurements, we use a sensing matrix which is constructed from rows of a Hadamard matrix of order $N=2^{14}$. Each row of the Hadamard matrix is permuted according to a predetermined random permutation.
The scene is composed of a photo printed on a paper and we capture the measurements of this photo.
Example results using different numbers of measurement are shown in Fig.~\ref{fig:Lena_real}. 
We also compare the five algorithms used in the simulation.
It can be seen that, similar to the simulation, our proposed algorithm provides best result when CSr is small. Especially, at CSr = 0.05 and 0.1, our algorithm can present many details of the face, for example, the left eye of ``Lena".
D-AMP introduces some ``blob" noise because the BM3D denoising approach is used.
Though NLR-CS can provide good results at CSr = 0.05 and 0.1, it introduces some unpleasant artifacts when CSr is from 0.15 to 0.25.
We also tried the algorithm (sHM) proposed in~\cite{Yuan14TSP}, where a Bayesian model is developed to investigate the tree-structure in wavelet. Surprisingly, sHM now works better than TVAL3 and GAP, mainly due to the following two reasons. Firstly, the tree structure in wavelet helps the reconstruction and secondly, the Bayesian framework developed in~\cite{Yuan14TSP} is very robust to noise; it infers noise from the measurements.   

\begin{figure}[htbp!]
	\centering
	\includegraphics[width=0.5\textwidth]{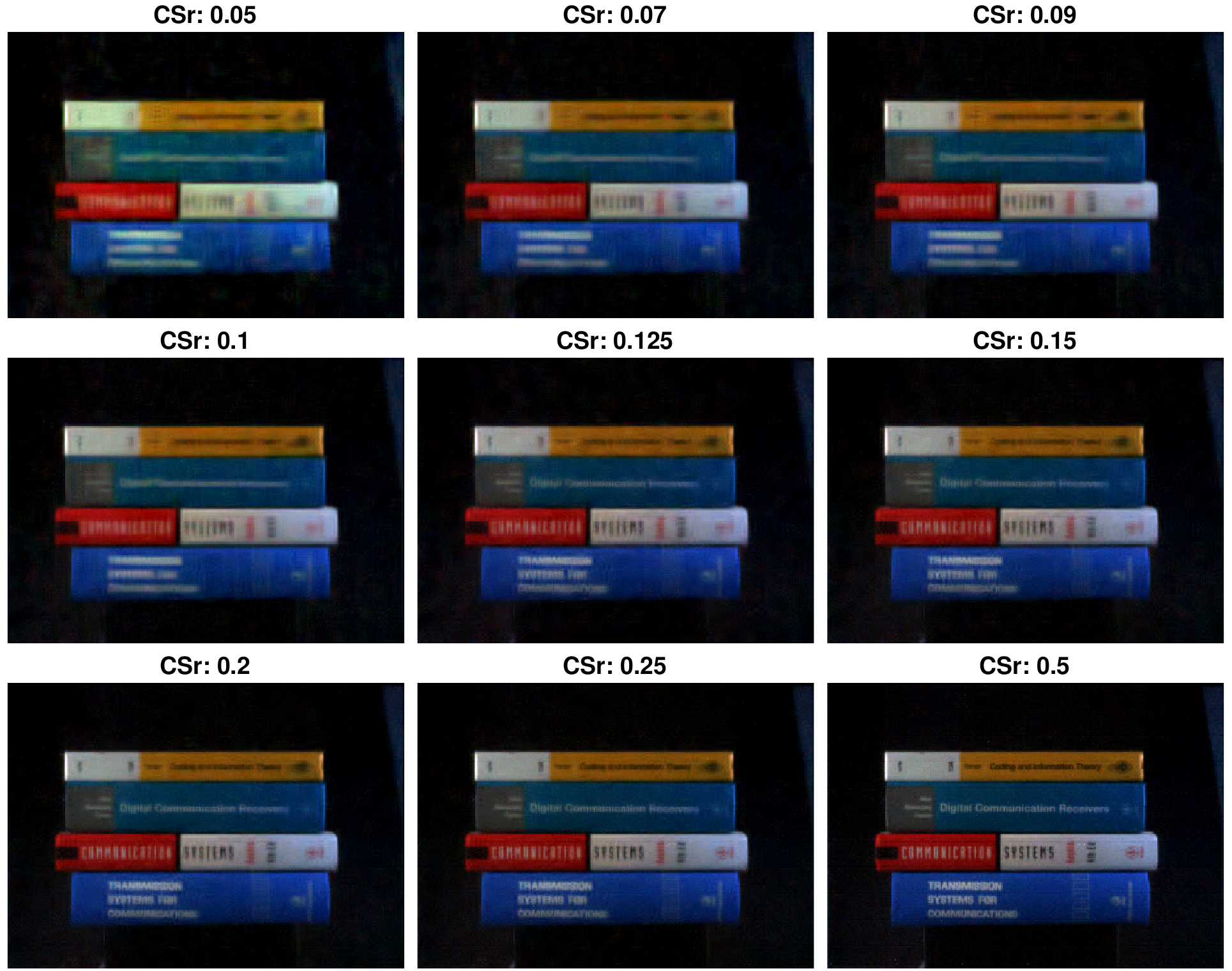}
	\vspace{-4mm}
	\caption{Real data: reconstruction results at different CSr with the proposed algorithm. The image is of size $217\times 302 \times 3$.}
	\label{fig:Books_real}
\end{figure}
\subsection{RGB Images}
Next we consider the RGB images captured by the RGB sensor, and now the resolution is $217\times 302\times 3$. The sensing matrix is constructed from rows of a Hadamard matrix of order $N=2^{16}$ and the first $65534$ elements are used.
The scene is the real scene of four books as shown~\cite{Huang13ICIP,Jiang14APSIPA}.
The reconstruction result is shown in Figure~\ref{fig:Books_real} with diverse CSr.

Note that by using compressive measurements, we can save the sensors as well as the bandwidth. As stated earlier, we may progressively get better results by receiving more measurements.
One of the main usage of compressive sensing is to get features in limited data by using a small bandwidth. 
From the results in Figure~\ref{fig:Lena_real}, we may identity high quality features from the reconstructed image at CSr around 0.1.
If we want to get some details, for example, the book titles in Figure~\ref{fig:Books_real}, we may need CSr around 0.2.  
On the other hand, if we only need to identify that these are ``books" in Figure~\ref{fig:Books_real}, CSr at 0.05 may be sufficient.

\section{Conclusions}
\label{Sec:Con}
A novel compressive sensing reconstruction algorithm is developed via exploiting the low-rank property of overlapping patches. A general iteratively two-step framework for compressive sensing recovery is proposed. A denoising operator is used to update the estimate of the desired image (obtained by the projection of the measurements), which can be implemented by investigating the sparsity or low-rank property of the image patches.  
We develop a probabilistic regime by representing each patch via a Gaussian mixture model and impose low-rank on each Gaussian component to achieve the state-of-the-art compressive sensing reconstruction results, in particular when the measurement number is small.
Additionally, the proposed low-rank GMM algorithm degrades to the low-rank piecewise linear estimator if each patch is modeled by a single Gaussian model.
Extensive results on both simulation and real data demonstrate high performance of the proposed algorithm.

\bibliographystyle{IEEEtran}


\begin{thebibliography}{10}
	\providecommand{\url}[1]{#1}
	\csname url@samestyle\endcsname
	\providecommand{\newblock}{\relax}
	\providecommand{\bibinfo}[2]{#2}
	\providecommand{\BIBentrySTDinterwordspacing}{\spaceskip=0pt\relax}
	\providecommand{\BIBentryALTinterwordstretchfactor}{4}
	\providecommand{\BIBentryALTinterwordspacing}{\spaceskip=\fontdimen2\font plus
		\BIBentryALTinterwordstretchfactor\fontdimen3\font minus
		\fontdimen4\font\relax}
	\providecommand{\BIBforeignlanguage}[2]{{%
			\expandafter\ifx\csname l@#1\endcsname\relax
			\typeout{** WARNING: IEEEtran.bst: No hyphenation pattern has been}%
			\typeout{** loaded for the language `#1'. Using the pattern for}%
			\typeout{** the default language instead.}%
			\else
			\language=\csname l@#1\endcsname
			\fi
			#2}}
	\providecommand{\BIBdecl}{\relax}
	\BIBdecl
	
	\bibitem{Donoho06ITT}
	D.~L. Donoho, ``Compressed sensing,'' \emph{IEEE Transactions on Information
		Theory}, vol.~52, no.~4, pp. 1289--1306, April 2006.
	
	\bibitem{Candes06ITT}
	E.~J. Cand\`{e}s, J.~Romberg, and T.~Tao, ``Robust uncertainty principles:
	Exact signal reconstruction from highly incomplete frequency information,''
	\emph{IEEE Transactions on Information Theory}, vol.~52, no.~2, pp. 489--509,
	February 2006.
	
	\bibitem{Candes08SPM}
	E.~J. Cand\`{e}s and M.~B. Wakin, ``An introduction to compressive sampling,''
	\emph{IEEE Signal Processing Magazine}, vol.~25, no.~2, pp. 21--30, March
	2008.
	
	\bibitem{Baraniuk07SPM}
	R.~Baraniuk, ``Compressive sensing,'' \emph{IEEE Signal Processing Magazine},
	vol.~24, no.~4, pp. 118--121, July 2007.
	
	\bibitem{Huang13ICIP}
	G.~Huang, H.~Jiang, K.~Matthews, and P.~Wilford, ``Lensless imaging by
	compressive sensing,'' \emph{IEEE International Conference on Image
		Processing}, 2013.
	
	\bibitem{Yuan15Lensless}
	X.~Yuan, H.~Jiang, G.~Huang, and P.~Wilford, ``Lensless compressive imaging,''
	\emph{arXiv:1508.03498}, 2015.
	
	\bibitem{Patrick13OE}
	P.~Llull, X.~Liao, X.~Yuan, J.~Yang, D.~Kittle, L.~Carin, G.~Sapiro, and D.~J.
	Brady, ``Coded aperture compressive temporal imaging,'' \emph{Optics
		Express}, pp. 698--706, 2013.
	
	\bibitem{Yuan14CVPR}
	X.~Yuan, P.~Llull, X.~Liao, J.~Yang, G.~Sapiro, D.~J. Brady, and L.~Carin,
	``Low-cost compressive sensing for color video and depth,'' in \emph{IEEE
		Conference on Computer Vision and Pattern Recognition (CVPR)}, 2014.
	
	\bibitem{Yuan13ICIP}
	X.~Yuan, J.~Yang, X.~Liao, P.~Llull, G.~Sapiro, D.~J. Brady, and L.~Carin,
	``Adaptive temporal compressive sensing for video,'' \emph{IEEE International
		Conference on Image Processing}, pp. 1--4, 2013.
	
	\bibitem{Yuan15FiO}
	X.~Yuan and S.~Pang, ``Structured illumination temporal compressive
	microscopy,'' in \emph{Frontier in Optics (FiO)}, 2015.
	
	\bibitem{Stevens15ASCI}
	A.~Stevens, L.~Kovarik, P.~Abellan, X.~Yuan, L.~Carin, and N.~D. Browning,
	``Applying compressive sensing to tem video: A substantial framerate increase
	on any camera,'' \emph{Advanced Structural and Chemical Imaging}, 2015.
	
	\bibitem{Llull14COSI}
	P.~Llull, X.~Yuan, X.~Liao, J.~Yang, L.~Carin, G.~Sapiro, and D.~Brady,
	``Compressive extended depth of field using image space coding,'' in
	\emph{Computational Optical Sensing and Imaging (COSI)}, 2014, pp. 1--3.
	
	\bibitem{Wagadarikar08CASSI}
	A.~Wagadarikar, R.~John, R.~Willett, and D.~J. Brady, ``Single disperser design
	for coded aperture snapshot spectral imaging,'' \emph{Applied Optics},
	vol.~47, no.~10, pp. B44--B51, 2008.
	
	\bibitem{Yuan15JSTSP}
	X.~Yuan, T.-H. Tsai, R.~Zhu, P.~Llull, D.~J. Brady, and L.~Carin, ``Compressive
	hyperspectral imaging with side information,'' \emph{IEEE Journal of Selected
		Topics in Signal Processing}, vol.~9, no.~6, pp. 964--976, September 2015.
	
	\bibitem{Tsai15OL}
	T.-H. Tsai, P.~Llull, X.~Yuan, D.~J. Brady, and L.~Carin, ``Spectral-temporal
	compressive imaging,'' \emph{Optics Letters}, vol.~40, no.~17, pp.
	4054--4057, Sep 2015.
	
	\bibitem{Tsai15OE}
	T.-H. Tsai, X.~Yuan, and D.~J. Brady, ``Spatial light modulator based color
	polarization imaging,'' \emph{Optics Express}, vol.~23, no.~9, pp.
	11\,912--11\,926, May 2015.
	
	\bibitem{Chan08APL}
	W.~L. Chan, K.~Charan, D.~Takhar, K.~F. Kelly, R.~G. Baraniuk, and D.~M.
	Mittleman, ``A single-pixel terahertz imaging system based on compressed
	sensing,'' \emph{Applied Physics Letters}, vol.~93, no.~12, pp.
	121\,105--–121\,105--–3, 2008.
	
	\bibitem{Babacan11ICIP}
	S.~Babacan, M.~Luessi, L.~Spinoulas, A.~Katsaggelos, N.~Gopalsami, T.~Elmer,
	R.~Ahern, S.~Liao, and A.~Raptis, ``Compressive passive millimeter-wave
	imaging,'' \emph{International Conference on Image Processing}, pp.
	2705--2708, 2011.
	
	\bibitem{Yang14GMMonline}
	J.~Yang, X.~Liao, X.~Yuan, P.~Llull, D.~J. Brady, G.~Sapiro, and L.~Carin,
	``Compressive sensing by learning a {G}aussian mixture model from
	measurements,'' \emph{IEEE Transaction on Image Processing}, vol.~24, no.~1,
	pp. 106--119, January 2015.
	
	\bibitem{Jiang14APSIPA}
	H.~Jiang, G.~Huang, and P.~Wilford, ``Multi-view in lensless compressive
	imaging,'' \emph{APSIPA Transactions on Signal and Information Processing},
	vol.~3, no.~15, pp. 1--10, 2014.
	
	\bibitem{Figueiredo07GPSR}
	M.~A.~T. Figueiredo, R.~D. Nowak, and S.~J. Wright, ``Gradient projection for
	sparse reconstruction: Application to compressed sensing and other inverse
	problems,'' pp. 586--597, Dec. 2007.
	
	\bibitem{Figueiredo07MM}
	M.~A. Figueiredo, J.~M. Bioucas-Dias, and R.~D. Nowak,
	``Majorization–minimization algorithms for wavelet-based image
	restoration,'' \emph{IEEE Transactions on Image Processing}, vol.~16, no.~12,
	pp. 2980--2991, 2007.
	
	\bibitem{Yin08bregman}
	W.~Yin, S.~Osher, D.~Goldfarb, and J.~Darbon, ``Bregman iterative algorithms
	for $\ell_1$-minimization with applications to compressed sensing,''
	\emph{SIAM J. Imaging Sci}, pp. 143--168, 2008.
	
	\bibitem{Candes08L1}
	E.~Candes, M.~Wakin, and S.~Boyd, ``Enhancing sparsity by reweighted $\ell_1$
	minimization,'' \emph{Journal of Fourier Analysis and Applications}, vol.~14,
	no.~5, pp. 877--905, 2008.
	
	\bibitem{Tropp07ITT}
	J.~A. Tropp and A.~C. Gilbert, ``Signal recovery from random measurements via
	orthogonal matching pursuit,'' \emph{IEEE Transactions on Information
		Theory}, 2007.
	
	\bibitem{daubechies2010iteratively}
	I.~Daubechies, R.~DeVore, M.~Fornasier, and C.~S. G{\"u}nt{\"u}rk,
	``Iteratively reweighted least squares minimization for sparse recovery,''
	\emph{Communications on Pure and Applied Mathematics}, vol.~63, no.~1, pp.
	1--38, 2010.
	
	\bibitem{Bioucas-Dias2007TwIST}
	J.~Bioucas-Dias and M.~Figueiredo, ``A new {TwIST}: Two-step iterative
	shrinkage/thresholding algorithms for image restoration,'' \emph{IEEE
		Transactions on Image Processing}, vol.~16, no.~12, pp. 2992--3004, December
	2007.
	
	\bibitem{Ji08SPT}
	S.~Ji, Y.~Xue, and L.~Carin, ``Bayesian compressive sensing,'' \emph{IEEE
		Transactions on Signal Processing}, vol.~56, no.~6, pp. 2346--2356, June
	2008.
	
	\bibitem{He09SPT}
	L.~He and L.~Carin, ``Exploiting structure in wavelet-based bayesian
	compressive sensing,'' \emph{IEEE Transactions on Signal Processing},
	vol.~57, no.~9, pp. 3488--3497, September 2009.
	
	\bibitem{Li13COA}
	C.~Li, W.~Yin, H.~Jiang, and Y.~Zhang, ``An efficient augmented lagrangian
	method with applications to total variation minimization,''
	\emph{Computational Optimization and Applications}, vol.~56, no.~3, pp.
	507--530, 2013.
	
	\bibitem{Huang14TIP}
	Y.~Huang, J.~Paisley, Q.~Lin, X.~Ding, X.~Fu, and X.~Zhang, ``Bayesian
	nonparametric dictionary learning for compressed sensing {MRI},'' \emph{IEEE
		Transactions on Image Processing}, vol.~23, no.~12, pp. 5007--5019, December
	2014.
	
	\bibitem{Averbuch12SIAM}
	S.~D. A.~Averbuch and S.~Deutsch, ``Adaptive compressed image sensing using
	dictionaries,'' \emph{SIAM Journal on Imaging Sciences}, vol.~5, no.~1, pp.
	57--89, 2012.
	
	\bibitem{Dong14TIP}
	W.~Dong, G.~Shi, X.~Li, Y.~Ma, and F.~Huang, ``Compressive sensing via nonlocal
	low-rank regularization,'' \emph{IEEE Transactions on Image Processing},
	vol.~23, no.~8, pp. 3618--3632, 2014.
	
	\bibitem{Mertzler14Denoising}
	C.~A. Metzler, A.~Maleki, and R.~G. Baraniuk, ``From denoising to compressed
	sensing,'' \emph{arXiv:1406.4175}, 2014.
	
	\bibitem{Beck09IST}
	A.~Beck and M.~Teboulle, ``A fast iterative shrinkage-thresholding algorithm
	for linear inverse problems,'' \emph{SIAM J. Img. Sci.}, vol.~2, no.~1, pp.
	183--202, Mar. 2009.
	
	\bibitem{Liao14GAP}
	X.~Liao, H.~Li, and L.~Carin, ``Generalized alternating projection for
	weighted-$\ell_{2,1}$ minimization with applications to model-based
	compressive sensing,'' \emph{SIAM Journal on Imaging Sciences}, vol.~7,
	no.~2, pp. 797–--823, 2014.
	
	\bibitem{ADMM2011Boyd}
	S.~Boyd, N.~Parikh, E.~Chu, B.~Peleato, and J.~Eckstein, ``Distributed
	optimization and statistical learning via the alternating direction method of
	multipliers,'' \emph{Found. Trends Mach. Learn.}, vol.~3, no.~1, pp. 1--122,
	January 2011.
	
	\bibitem{Dabov07BM3D}
	K.~Dabov, A.~Foi, V.~Katkovnik, and K.~Egiazarian, ``Image denoising by sparse
	3d transform-domain collaborative filtering,'' \emph{IEEE Transactions on
		Image Processing}, vol.~16, no.~8, pp. 2080--2095, August 2007.
	
	\bibitem{Elad06TIP}
	M.~Elad and M.~Aharon, ``Image denoising via sparse and redundant
	representations over learned dictionaries,'' \emph{IEEE Transactions on Image
		Processing}, vol.~15, pp. 3736--–3745, December 2006.
	
	\bibitem{Yu11SPT}
	G.~Yu and G.~Sapiro, ``Statistical compressed sensing of {G}aussian mixture
	models,'' \emph{IEEE Transactions on Signal Processing}, vol.~59, no.~12, pp.
	5842--5858, 2011.
	
	\bibitem{Yu12IPT}
	G.~Yu, G.~Sapiro, and S.~Mallat, ``Solving inverse problems with piecewise
	linear estimators: From {G}aussian mixture models to structured sparsity,''
	\emph{IEEE Transactions on Image Processing}, 2012.
	
	\bibitem{Jiang15TSP}
	H.~Jiang, G.~Huang, P.~A. Wilford, and L.~Yu, ``Constrained and preconditioned
	stochastic gradient method,'' \emph{IEEE Transactions on Signal Processing},
	vol.~63, no.~10, pp. 2678--2691, 2015.
	
	\bibitem{Yang14GMM}
	J.~Yang, X.~Yuan, X.~Liao, P.~Llull, G.~Sapiro, D.~J. Brady, and L.~Carin,
	``Video compressive sensing using {G}aussian mixture models,'' \emph{IEEE
		Transaction on Image Processing}, vol.~23, no.~11, pp. 4863--4878, November
	2014.
	
	\bibitem{Chen10SPT}
	M.~Chen, J.~Silva, J.~Paisley, C.~Wang, D.~Dunson, and L.~Carin, ``Compressive
	sensing on manifolds using a nonparametric mixture of factor analyzers:
	Algorithm and performance bounds,'' \emph{IEEE Transactions on Signal
		Processing}, vol.~58, no.~12, pp. 6140--6155, December 2010.
	
	\bibitem{Gut2009}
	A.~Gut, \emph{An Intermediate Course in Probability}.\hskip 1em plus 0.5em
	minus 0.4em\relax Springer, 2009.
	
	\bibitem{Cai10SVT}
	J.-F. Cai, E.~J. Cand\`{e}s, and Z.~Shen, ``A singular value thresholding
	algorithm for matrix completion,'' \emph{SIAM J. on Optimization}, vol.~20,
	no.~4, pp. 1956--1982, Mar. 2010.
	
	\bibitem{Romberg08SPM}
	J.~Romberg, ``Imaging via compressive sampling,'' \emph{IEEE Signal Processing
		Magazine}, vol.~25, no.~2, pp. 14--20, 2008.
	
	\bibitem{Jiang12Inverse}
	H.~Jiang, W.~Deng, and Z.~Shen, ``Surveillance video processing using
	compressive sensing,'' \emph{Inverse Problems and Imaging}, vol.~5, no.~2,
	pp. 201--214, 2012.
	
	\bibitem{Yuan14TSP}
	X.~Yuan, V.~Rao, S.~Han, and L.~Carin, ``Hierarchical infinite divisibility for
	multiscale shrinkage,'' \emph{IEEE Transactions on Signal Processing},
	vol.~62, no.~17, pp. 4363--4374, Sep. 1 2014.
	
\end{thebibliography}

\end{document}